\documentclass{article}

\usepackage{arxiv}
\usepackage[USenglish]{babel}
\usepackage[utf8]{inputenc} 
\usepackage[T1]{fontenc}    
\usepackage{hyperref}       
\usepackage{url}            
\usepackage{booktabs}       
\usepackage{amsfonts}       
\usepackage{nicefrac}       
\usepackage{microtype}      
\usepackage{lipsum}		
\usepackage{graphicx} 
\usepackage{doi}
\usepackage{bm}
\usepackage{amsmath}
\usepackage{amssymb,MnSymbol, multirow}
\usepackage{longtable} 
\usepackage{array}
\usepackage{rotating}
\usepackage{color, colortbl}
\definecolor{Gray}{gray}{0.9}
\usepackage{titlesec}
\usepackage[nolist]{acronym}

\title{Graph Reinforcement Learning for Power Grids: A Comprehensive Survey}

\date{} 				
\author{
    \textbf{Mohamed~Hassouna}\textsuperscript{1,2,}\thanks{Both authors contributed equally.\\ Email addresses: mohamed.hassouna@iee.fraunhofer.de, clara.juliane.holzhueter@iee.fraunhofer.de, pawel.lytaev@uni-kassel.de, josephine.thomas@uni-kassel.de, bsick@uni-kassel.de, christoph.scholz@iee.fraunhofer.de} \and
    \textbf{Clara~Holzhüter}\textsuperscript{1,2,}\footnotemark[1]\and
    Pawel~Lytaev\textsuperscript{3} \and
    Josephine~Thomas\textsuperscript{4} \and
    Bernhard~Sick\textsuperscript{2} \and
    Christoph~Scholz\textsuperscript{1,2} \and
\\
\textsuperscript{1} Fraunhofer Institute for Energy Economics and Energy System Technology (IEE)\\
\textsuperscript{2} Intelligent Embedded Systems, University of Kassel \\
\textsuperscript{3} Department of Sustainable Electrical Energy Systems, University of Kassel\\
\textsuperscript{4} Machine Learning, University of Greifswald
\\
\\
}

\renewcommand{\shorttitle}{Graph Reinforcement Learning for Power Grids: A Comprehensive Survey}

\hypersetup{
pdftitle={Graph Reinforcement Learning in Power Grids: A Survey},
pdfsubject={q-bio.NC, q-bio.QM},
pdfauthor={David S.~Hippocampus, Elias D.~Striatum},
pdfkeywords={First keyword, Second keyword, More},
}

\begin{document}
\maketitle

\begin{abstract}
    
The increasing share of renewable energy and distributed electricity generation requires the development of deep learning approaches to address the lack of flexibility inherent in traditional power grid methods. In this context, Graph Neural Networks are a promising solution due to their ability to learn from graph-structured data. Combined with Reinforcement Learning, they can be used as control approaches to determine remedial actions. This review analyses how Graph Reinforcement Learning can improve representation learning and decision-making in power grid applications, particularly transmission and distribution grids. We analyze the reviewed approaches in terms of the graph structure, the Graph Neural Network architecture, and the Reinforcement Learning approach. Although Graph Reinforcement Learning has demonstrated adaptability to unpredictable events and noisy data, its current stage is primarily proof-of-concept, and it is not yet deployable to real-world applications. We highlight the open challenges and limitations for real-world applications.

\end{abstract}


\section{Introduction}

The role of electrical grid operators, both for transmission and distribution grids, is to ensure cost-efficient system availability at all times. However, power systems worldwide are undergoing a paradigm shift driven by the need for CO2 neutrality. The integration of distributed renewable generation and additional load demand due to heating and traffic sector electrification introduce complexities that traditional power system operation struggles to cope with. These trends require advanced methods for optimal operation \cite{marot2021learning,kelly_reinforcement_2020}. The ongoing energy transition also impacts other stakeholders, including energy market participants. They need to adapt to the decentralized structure and new market players such as \ac{EV} charging operators. Additionally, the ongoing digitization and build-up of communication systems transform the classical power system into a \ac{CPES} \cite{Steinbrink2018}. These new challenges introduce a new layer of complexity to power grid operation.

Traditionally, grid operation has mostly relied on optimization approaches for \ac{OPF} problems \cite{Bienstock2022, CAPITANESCU2011}. However, due to the non-linear and non-convex nature of \acp{OPF}, these approaches struggle to scale to real power grids such that exact results cannot be obtained in a reasonable time \cite{srivastava_voltage_2023, CAPITANESCU2016}. Therefore, relaxation techniques are used to reduce the complexity \cite{Molzahn2019, bienstock2014, coffrin2013}. However, they can produce imprecise results and cannot guarantee optimality, which casts doubt on their effectiveness in practical applications.

Furthermore, noisy or missing measurements cannot be reliably handled by classical approaches as they often assume ideal conditions and typically do not incorporate statistical models to distinguish between signal and noise \cite{liu2020noisy}. As a result, even small perturbations in the input can strongly affect the output. Therefore, practitioners are exploring deep learning solutions \cite{L2RPN,L2RPN_trust,chen2025physics} for power flow problems. Such solutions are promising alternatives to classical approaches, addressing the challenges of time criticality, scalability, and reliability of results. 

\ac{DRL} techniques can identify and exploit underutilized flexibility in power grids, which is often overlooked by traditional methods and human operators \cite{kelly_reinforcement_2020}. By learning from interactions with the grid environment, \ac{DRL} agents can dynamically adapt to changing conditions and unforeseen events, which can potentially prevent cascading failures and blackouts \cite{marot2021learning, Donnot2017}. Furthermore, their ability to consider long-term horizons aligns with the dynamic nature of power grids \cite{Viebahn2022}. However, the development and training of \ac{DRL} agents requires extensive simulations, as direct interaction with the physical grid is impractical. These simulations often need to abstract from reality and rarely use real data, leading to challenges in transferring the solutions back to real-world applications \cite{sim2real}. Furthermore, the large combinatorial action spaces in power grids hinder the application of \ac{DRL} \cite{Dulac_Arnold2021}, highlighting the need for handcrafted action spaces and other reduction techniques \cite{binbinchen, lehna2022reinforcement, lehna2024hugo}. Despite these challenges, the potential for \ac{RL} in power grid management is significant, particularly as power systems strive to meet decarbonization goals \cite{Prostejovsky2019, Marot2022ai}. The aim is not to replace human operators but to provide them with \ac{RL}-driven action recommendations \cite{Viebahn2022, Marot_Perspectives}. Despite impressive proof-of-concept results, DRL research for grid control is still in its early stages, and significant gaps remain to be filled before deployment.

Power grids can be naturally modeled as a graph, in which nodes and edges correspond to grid elements and the connections between them \cite{Viebahn2022}. Since \acp{GNN} are designed explicitly for such graph-structured data, they are highly suitable for modeling interdependencies in power systems \cite{liao_review_2021}. They can capture relationships between elements and enable an effective feature extraction from the grid. While standard, i.e., non-graph-based, neural networks struggle to produce accurate results when the grid's topology changes, \acp{GNN} are more robust to modifications of the graph structure. This is an advantage, as several grid actions, such as bus-bar splitting, can transform a single node into two distinct nodes (or combine two nodes into one). This is uncommon in other types of networks \cite{Donon2020} and thus requires tailored methods. 
Similarly, long-range dependencies and the rapid propagation speed of electricity pose unique challenges. As GNN research is strongly driven by standard benchmark datasets such as citation networks, molecules, or social networks, most architectures are not entirely suited for the graph structure and properties of power grids. Therefore, the design of tailored GNN architectures, including the generalization across different topologies and grids, still requires extensive research \cite{Ringsquandl2021}. With regard to the practical applicability of \acp{GNN}, the interpolation capabilities of \acp{GNN} are crucial in reconstructing missing information and smoothing out noisy measurements \cite{Kuppannagari2021}. This takes into account the fact that sensors can experience connectivity problems, resulting in incomplete or unreliable data. 

The combination of \acp{GNN} and \ac{RL} represents a synergy that exploits the strengths of both paradigms. \acp{GNN} provide a powerful tool for feature extraction in graph-structured data, enhancing the \ac{RL} agent's understanding of the complex relationships within power grids. As pointed out by \cite{munikoti_challenges_2023}, the performance of \ac{RL} agents heavily depends on the state encoding, and \acp{GNN} are much better encoders for graph-structured environments. Incorporating them into \ac{RL} has the potential for more informed decision-making, better adaptability to changing network conditions and noise, and improved generalization across different scenarios and topologies.

\paragraph{Existing Works}
\label{rel_work}
This survey provides a comprehensive overview of existing GRL approaches for grid applications, filling a significant gap in the literature. While the \ac{GRL} literature on power systems is growing, it remains fragmented across \ac{GNN}, \ac{RL}, \ac{GRL}, and other deep learning approaches for power grids. This highlights the need for a comprehensive comparison of existing approaches to inform future work. 

Table~\ref{table:existing_works} lists the works discussed in more detail below, along with the methodologies and use cases they address. For clarity, the table only includes works that address at least one methodology and one use case. Broader \ac{GNN} and \ac{RL} surveys are not included.

From a methodological perspective, several broader surveys on \acp{GNN} cover various methodologies and applications. For example, \cite{zhou2020graph}, \cite{thomas2022graph}, and \cite{wu2020comprehensive} provide overviews of common architectures. There are also various reviews with a specific focus, such as \cite{skarding2021foundations}, which focuses on dynamic graphs that evolve over time, and \cite{wu2022graph} examines GNNs in recommender systems. For the power grids use case, two reviews on specialized \acp{GNN} are available. \cite{liao_review_2021} highlight the superior performance of \acp{GNN} over standard neural networks, particularly in fault analysis, time series prediction, power flow calculation, and data generation. The second paper, \cite{li2023graph}, focuses on the power flow problem and the respective benefits and challenges. Neither paper covers the combination with \ac{RL} or other sequential decision-making algorithms.

Similarly in \ac{RL}, general reviews include \cite{arulkumaran2017deep}, \cite{garcia2015comprehensive}, and \cite{zhu2023transfer}. \cite{zhang2019deep} explores \ac{DRL} for energy systems, focusing on problems such as demand response, electricity markets, and operational control. Specialized reviews, such as \cite{vazquez-canteli_reinforcement_2019}, focus on demand response in smart grids but do not cover \acp{GNN}, which is a more recent development compared to traditional \ac{RL} approaches. Furthermore, a recent study analyzes \ac{RL} approaches for power grid control, applied to the Grid2op framework \cite{van2025survey}.

Literature reviews combining \ac{GNN} and \ac{RL} are rare. \cite{munikoti_challenges_2023} survey 80 relevant papers, categorizing them into \ac{DRL}-enhancing \acp{GNN} and \acp{GNN}-enhancing \ac{DRL}. The former includes \ac{DRL} for architecture search and improving \ac{GNN} explainability, while the latter covers the use of \ac{GNN} in \ac{DRL}, which is more closely related to our work. They explore areas like combinatorial optimization and transportation, but exclude energy applications.

\cite{fathinezhad_graph_2023} survey \ac{GRL} approaches with a focus on the methodology of \acp{GNN} and \ac{RL}, especially in multi-agent settings where \acp{GNN} facilitate agent communication. They primarily explore how graphs and \ac{RL} interact, while we focus on using \acp{GNN} as feature extractors for graph-structured power grid data. Although they briefly mention an energy-related application, it is not analyzed in detail as their review does not emphasize application-specific approaches.

Similarly, the survey presented by \cite{nie2023reinforcement} examines \ac{GRL} methodologically, detailing how \ac{RL} can enhance \acp{GNN} and address graph problems. They cover various transportation and medical research applications but do not address energy-related use cases.

Several surveys focus on specific aspects of \ac{GRL}: \cite{mazyavkina_reinforcement_2021} examines \acp{GNN} for representation learning in combinatorial optimization within \ac{RL}. \cite{mendonca2019graph} detail how graph algorithms can enhance \ac{RL} through action abstraction, while \cite{pateria2021hierarchical} explore hierarchical \ac{RL} with graph-based approaches for discovering subtasks. None of these surveys addresses power grids.

Further related works analyze specific power grid problems such as voltage control (e.g. \cite{srivastava_voltage_2023} \cite{murray_voltage_2021}). However, they mostly review traditional optimization and heuristics. To our knowledge, no comprehensive overview of \ac{GRL} in power grids exists. Therefore, this work aims to fill this gap. 

\begin{table*}[t]
\label{table:existing_works}
\centering
\renewcommand{\arraystretch}{1.2}
\begin{tabular}{|l|c|c|c|c|c|c|}
\hline
\multirow{2}{*}{Reference} & \multicolumn{3}{c|}{Methodology} & \multicolumn{3}{c|}{Grid Use Case} \\
\cline{2-7}
 & GNNs & RL & GRL & Transmission& Distribution& Other\\
\hline
\hline
\cite{liao_review_2021} (Liao et al., 2021) & $\checkmark$ &  &  & $\checkmark$ & $\checkmark$ &  \\
\hline
\cite{li2023graph} (Li et al., 2023) & $\checkmark$ & & & $\checkmark$ & $\checkmark$ &   \\
\hline
\cite{zhang2019deep} (Zhang et al., 2019) &  & $\checkmark$ &  & $\checkmark$ & $\checkmark$ & $\checkmark$ \\
\hline
\cite{vazquez-canteli_reinforcement_2019} (Vázquez-Canteli et al., 2019) &  & $\checkmark$ &  & $\checkmark$ & $\checkmark$ & $\checkmark$ \\
\hline
\cite{van2025survey} (van der Sar et al., 2025) &  & $\checkmark$ & ($\checkmark$) & $\checkmark$ &  &  \\
\hline
\cite{fathinezhad_graph_2023} (Fathinezhad et al., 2023) & $\checkmark$ & $\checkmark$ & ($\checkmark$) &  &  & ($\checkmark$) \\
\hline
\hline
\textbf{Ours} & $\checkmark$ & $\checkmark$ &$\checkmark$ & $\checkmark$ & $\checkmark$ & $\checkmark$\\
\hline
\end{tabular}
\caption{\textbf{Overview of existing surveys} that address at least one power grid use case and one methodological aspect.}
\end{table*}
\paragraph{Contribution and Structure}
Addressing the fragmented existing literature on \ac{GRL} approaches for power grids mentioned in the previous section, the core contribution of this survey is to provide the first comprehensive analysis of \ac{GRL} approaches designed for power grid applications, covering transmission and distribution grids. We also investigate other power grid applications, such as the energy market, communication networks within power systems, and \ac{EV} charging management, if they consider the underlying power grid in their approach. We focus on approaches that utilize \acp{GNN} to capture the graph-structured state space of power grids while applying \ac{DRL} for sequential decision-making.

The main contributions are as follows:
\begin{itemize}
    \item \textbf{Comprehensive review:} We are the first to provide a comprehensive analysis of existing \ac{GRL} approaches for power grid use cases. We provide an overview of the applied \ac{GRL} techniques, including states, actions, and rewards, and analyze the proposed \ac{GNN} architecture in detail. We highlight the commonalities and differences between the analyzed methods and identify the most common approaches.
    \item \textbf{Categorization}: We categorize the approaches based on the specific scenarios they address. Our analysis focuses on applications in distribution and transmission grids. Within the distribution grid, we differentiate between regular voltage control and emergency situations. For the transmission grid, our focus is on topology control and the relevant frameworks.
     \item \textbf{Future Directions:} We highlight crucial aspects for the application of \ac{GRL} in real-world scenarios and investigate limitations and open challenges of the proposed approaches.
\end{itemize}

The papers we analyze were published between 2020 and May 2025, as \ac{GRL} is a relatively new field. Our selection includes those papers that explicitly address power grids, including the underlying grid structure. 
To maintain a focused and in-depth analysis of GRL applied to large-scale grid operation, we deliberately limit our primary scope to transmission and distribution systems. Although GRL approaches exist for microgrids, the distinct operational and modeling complexities of these systems (e.g., islanding and distributed control) fall outside the scope of the present review. Therefore, we only include approaches for zoned grids and microgrids if they focus on the use case of operational control. 

This review is structured as follows: In Ch.~\ref{chap:basics}, we present the basics of transmission and distribution grids, \acp{GNN} and \ac{RL}. In the main part of this review, we discuss the presented methods in detail and categorize them by the use case they address. This part begins with approaches for common problems in transmission grids (Ch.~\ref {chap:TG}) and continues with applications in distribution grids (Ch.~\ref {chap:DG}). Then, in Ch.~\ref{chap:other}, we highlight several relevant papers catering to related use cases, such as \ac{EV} charging. Finally, we give an overall conclusion and overview of key challenges and future directions.
\section{Fundamentals: Power Grids, Graph Neural Networks and Reinforcement Learning}
\label{chap:basics}

\subsection{Power Grids}

Power grids are essential to modern society and a crucial part of today's infrastructure. In the face of the energy transition, power system engineers encounter challenges in all aspects of the grid.
Their purpose is to transport electricity from generation units to consumers, who are typically not located in the same area. Traditionally, generation has been centralized, for example, at fossil-fueled or nuclear power plants, resulting in a unidirectional power flow from generation to consumption. With the worldwide expansion of renewable energy, generation units are spread across the power system, resulting in a decentralized structure and a bi-directional power flow. With the electrification of the transportation and heating sectors, the consumption side of power grids is also undergoing a significant shift. This sector coupling not only increases overall consumption but also introduces new patterns of simultaneous demand. For instance, during winter, multiple heat pumps may operate concurrently across the network, which can significantly impact its load.

\begin{figure*}
	\centering
	\includegraphics[width=0.7\textwidth]{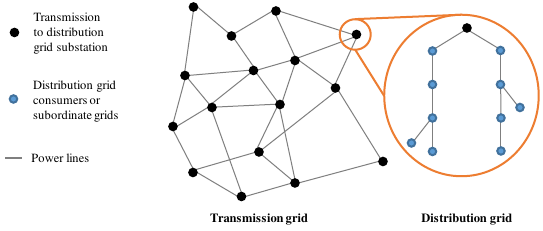}
	\caption{Visualization of the power grid structure with transmission and distribution level.}
 \label{fig:powergrids}
\end{figure*}

Power grids are typically divided into two levels: transmission and distribution. These levels are split by substations (see Fig.~\ref{fig:powergrids}) and differ in voltage levels, purpose, and characteristics. We elaborate on both levels in the following subsections.

\subsubsection{Transmission grids.}

The purpose of transmission grids is to transport large amounts of electricity over long distances that vary from a few hundred kilometers to a few thousand \cite{LiQiao2010}. Transmission grid operation aims to achieve at least N-1 secure operation. This means that, if one asset in the grid fails, the remaining grid is in a safe state. Therefore, transmission grids are typically built in meshed structures with redundancies installed, e.g., multiple transformers and busbars at substations \cite{Gui2024}. The structure of transmission grids yields a highly complex system that necessitates solving large-scale optimization problems.

One measure to prevent high grid congestion, such as line overloading, is re-dispatch, which refers to changing generator injections \cite{Linnemann2011}. Since generation and consumption in a power system must be balanced at all times, changing the set point of one generator set must result in the corresponding adjustment of another. This can lead to renewable generators being shut down and fossil fuel generators ramping up, which is undesirable in terms of both cost and CO2 neutrality. Therefore, other means of flexibility are being explored, such as topology control. By controlling the switching state in substations, the grid's topology can be modified, helping to reduce or even eliminate the need for re-dispatch. Optimizing the topology is a challenge in itself, as it results in a Mixed-Integer Non-Linear Problem. Here, deep learning solutions such as \ac{RL} can help \cite{L2RPN}.

\subsubsection{Distribution grids.}

Distribution grids cover smaller areas, such as villages, and often do not follow the N-1 safety criterion due to the lower impact of asset failures \cite{Escobar2021}.  They are undergoing a major shift due to the transformation from unidirectional power flows to bidirectional power flows coming from distributed generation.

Voltage volatility is higher due to time-varying generation and consumption, particularly with the increased adoption of photovoltaic generation. This can lead to voltage fluctuations. Traditional voltage control uses regulated transformers, shunt capacitors, and voltage regulators \cite{srivastava_voltage_2023}. With digitalization, voltage control options are expanding to include inverter-based technologies such as smart PV inverters \cite{HOWLADER2020pvactivepowercontrol}, vehicle-to-grid systems \cite{GONZALEZ2021activeEV}, and stationary batteries \cite{Stecca2020batteryreview}. These are flexibilities that grid operators must tap to maintain the system in a secure state.

Additionally, given the number of distribution grids, optimized control strategies need scalable and replicable solutions \cite{GONZALEZ2021activeEV}.

\subsubsection{Grid models.}
In research, synthetic grid models are widely used as a standard benchmark for evaluating new approaches and comparing results. 
Examples in literature include \cite{PGLib2021, MeineckeSimbench2020, MeineckeReview2020, cigre2014}. In the transmission and distribution grid use cases mentioned above, the IEEE test cases are the most commonly used benchmarks, available in several power system calculation tools, such as MATPOWER \cite{matpower}, pandapower \cite{pandapower}, and DIgSILENT PowerFactory \cite{powerfactory}. Although they comprise grid topology data, line connectivity, generator and load placements, line impedances, and capacity limits, these benchmark cases are simplified representations of power systems that do not fully capture their complexities and challenges. They are mostly simple bus branch models with loads and generators connected to the buses. More sophisticated assets, such as inverter-based generator controls, transformer tap changers, and shunt elements, are not commonly found in such grid models but are prevalent in real power grid operations. Moreover, these test cases often lack critical operational conditions such as fluctuating demand, renewable generation variability, or network contingencies like line faults. Therefore, methods benchmarked on such synthetic grids must be treated with caution, as their application in practice requires further consideration.
\subsection{Graph Neural Networks}
\begin{figure*}[h]
	\centering
	\includegraphics[width=\textwidth]{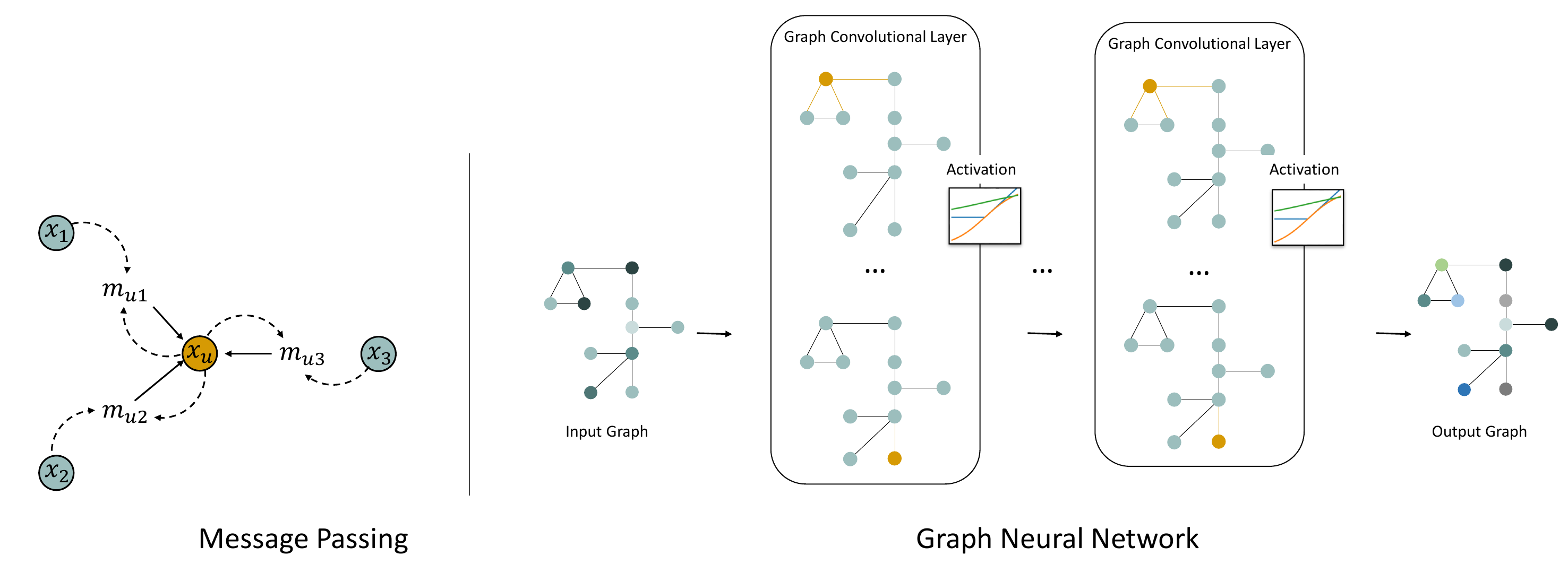}
	\caption{\textbf{Left}: Visualization of the general message passing scheme in \acp{GNN} (modeled after \cite{bronstein2021geometric}) - The target node (orange) receives messages $m_ui$ from its neighbors and aggregates them. The messages can be constructed from the information of both the target and neighboring node, depending on the message passing scheme. \textbf{Right}: Illustration of a \ac{GNN} (modeled after \cite{wu2020comprehensive}) - The graph is input to the \ac{GNN} layers, which compute node embeddings based on the messages from neighboring nodes. As indicated in orange, this is done for each node in the graph. After all embeddings are computed, an activation function is applied. This is repeated for a given number of layers. In the end, the \ac{GNN} outputs a graph with new node embeddings from which a prediction can be made. }
 \label{fig:hierarchy}
\end{figure*}

\acp{GNN} are designed to extract information from graph-structured data by applying multiple layers of graph convolutions. They can be interpreted as a generalization of \acp{CNN} to non-Euclidean structured data. The general idea is to combine information from local regions of the inputs in a learnable way and to grow these local regions from layer to layer. In this way, \ac{CNN} layers learn increasingly abstract features from the input data. \acp{CNN} perform exceptionally well on grid-structured data, such as images. However, many real-world phenomena involve relationships or complex dependencies that cannot be represented as regular grid structures without losing information. For example, in an image, every node (pixel) has the same number of neighbors, but in power grids, not every component is connected to the same number of power lines. 

Graphs consist of an unordered set of nodes and edges, where the edges define the neighborhood of a node. Graphs can, therefore, be used to model complex relationships, such as neighborhoods of arbitrary size or multiple types of edges or nodes. They can have attributes that describe properties of nodes and edges, such as node features or the strength of an edge. This makes graphs a flexible model for many real-world applications, such as power grids. 
Additionally, feed-forward neural networks operating on Euclidean data typically treat nodes as independent samples. This means that they neglect the relationships between nodes or stack them unsystematically into a vector. \acp{GNN}, on the contrary, make use of the information about node connectivity. They can solve all common learning tasks, i.e., classification, regression, and clustering, for entire graphs and at node- or edge-level. 
\subsubsection{Message Passing}
\label{subsubsec:messagepassing}
\acp{GNN} update the embeddings of the graph nodes by repeatedly aggregating information of their neighborhoods in a learnable way. This general scheme is known as message passing, where each node updates its embedding based on the messages it receives from its neighbors. The representation $\bm{h}_u'$ of a target node $u$ generated by a general message passing layer is computed as:
\begin{equation}
\label{equ:msgpassing}
\bm{h}_u' = \sigma\left( \bm{h}_u,  \bigoplus\limits_{v \in \mathcal{N}(u)}\psi (\bm{h}_u, \bm{h}_v)\right),
\end{equation}
where $\psi$ corresponds to a message function equipped with learnable weights that compute a message between node $u$ and its neighbor $v$. $\bm{h}_u$ and $\bm{h}_v$ are the respective node embeddings. In the first layer, these embeddings are simply the initial node features. $\mathcal{N}(u)$ is the neighborhood of node $u$, $\bigoplus$ refers to an aggregation function that defines how messages are passed \cite{book_bronstein_2021}, and $\sigma$ is an activation function. There exist various implementations of message passing layers; the scheme shown above is the most general one \cite{book_bronstein_2021}. This section describes the message passing schemes most frequently used among the analyzed approaches.

 Message passing in \acp{GNN} is linked to how electrical quantities are coupled in a power grid. The neighbour aggregation reflects the way buses in a power grid are electrically linked via the nodal admittance matrix. Stacking layers captures multi‑hop electrical influence, similar to a few iterations of simple iterative power‑flow methods (e.g., Gauss-Seidel method for meshed grids), where voltage and flow updates propagate along the network until nodal balances are satisfied. This provides a natural, physics-aligned inductive bias for learning on power grids.
\paragraph{Spatial Graph Convolution.} A simple form of message passing is the spatial graph convolution. Here, messages are the embeddings
of the neighboring nodes transformed using a learnable weight matrix. The aggregation corresponds to the summation operation. \cite{morris2019weisfeiler}, for example, propose such an intuitive formulation: 
\begin{equation}
\label{eq:morris}
    \bm{h}_u'= \sigma(\bm{W}\bm{h}_u + \sum_{v\in\mathcal{N}(u)} \bm{W} \bm{h}_v)
\end{equation}
with $\bm{W}$ being learnable weight matrices, also referred to as filters. Typically, the weights are shared across all nodes and neighbors. This concept follows the parameter sharing approach in \acp{CNN}.\\
\paragraph{GraphSage.}\label{graph_sage} The architecture proposed by \cite{hamilton2017inductive} is a special case of spatial graph convolution based on sampling. Instead of aggregating the entire neighborhood of a node in each layer, a fixed number of neighbors of the target node are randomly sampled. The neighbors are aggregated using a permutation-invariant function, such as the mean or maximum. GraphSage is trained in an unsupervised manner using a special loss function. It consists of two terms: one that enforces nodes close in the input graph have similar embeddings, and the other that pushes apart the embeddings of two nodes that are far apart in the graph.
\paragraph{Graph Attention Network}
A method commonly used to improve the performance of a \ac{GNN} model is to equip the graph convolution with an attention mechanism. Such layers are a special case of general message passing \cite{book_bronstein_2021} where attention coefficients are learned for each connected pair of nodes. They are computed from the features of the neighboring nodes and the target node and determine the influence a neighbor has on the target node. \cite{velivckovic2018graph} defines such an attention convolution that would extend Eq.~\ref{eq:morris} to:
\begin{equation}
\label{equ:att_conv}
    	\bm{h}_u' = \sigma (\bm{W}\bm{h}_u + \sum_{v\in\mathcal{N}(u)} \alpha_{u,v}\bm{W} \bm{h}_v)
\end{equation}
where $W$ refers to a learnable weight matrix, $\alpha_{u,v}$ refers to the attention coefficient for node $v$ in the neighbourhood aggregation of node $u$ indicating the importance of node $u$ to node  $v$. It is computed as: 

\begin{equation}
\label{equ:att_weights}
    	\bm{\alpha}_{u,v} := \text{softmax}_v \left(\sigma ( \bm{a}^T  \lbrack \bm{W} \bm{h}_u || \bm{W} \bm{h}_v \rbrack \right )
\end{equation}
with $\bm{a}$ and $\bm{W}$ being weight vector and matrix respectively. $\sigma$ again refers to an activation function, \cite{velivckovic2018graph} for example use ReLu. The $||$ corresponds to the concatenation operation, so the transformed features of both nodes are concatenated before the attention mechanism $\bm{a}$ is applied. The coefficients are normalized using softmax to make them comparable across the neighbors of the target node. 

\paragraph{Spectral Graph Convolution.}
Besides the aforementioned spatial \ac{GNN} layers, \acp{GNN} can be formulated using spectral theory, which refers to the study of the properties of linear operators. Similar to signal processing, where a signal can be decomposed into sine and cosine functions by Fourier decomposition, a graph signal $\bm{x}$ (a scalar for each node) can be transformed into the spectral domain by the graph Fourier transform $\mathop{F}$ and back with its inverse. Convolution in the spectral domain results in an element-wise multiplication, after which, the convolved signal is transformed back into the graph domain:
\begin{align}
\label{spectral_conv}
    \bm{g}*\bm{x} = \mathop{F^{-1}}(\mathop{F}(\bm{g})\mathop{F}(\bm{x}))=\bm{U}(\bm{U}^T\bm{g}\bm{U}^T\bm{x})
\end{align}
where $\bm{U}$ is the matrix of eigenvectors of the normalized graph Laplacian $\bm{L} = \bm{I}-\bm{D}^{-\frac{1}{2}}\bm{A}\bm{D}^{-\frac{1}{2}} $ and is determined by the eigendecomposition $\bm{L} = \bm{U}\bm{\Lambda}\bm{U}^T$. $\bm{U}^T\bm{g}$ is the filter in the spectral domain. Since the normalized graph Laplacian $\bm{L}$ is composed of the degree matrix $\bm{D}$ and adjacency matrix $\bm{A}$, intuitively the eigenvectors and eigenvalues indicate the main directions of information diffusion through the graph. A first-order approximation of the spectral graph convolution has been proposed by \cite{kipf2016semi}: 
\begin{equation}
\label{equ:spec_approx}
    	\bm{H}' = \sigma (\bm{D}^{-\frac{1}{2}}A\bm{D}^{-\frac{1}{2}}HW)
\end{equation}
where $H$ corresponds to the node feature matrix and W is a learnable weight matrix.

While spatial and spectral formulations of \acp{GNN} are equivalent, spatial \acp{GNN} are more commonly used in practice due to the high computational cost of spectral \acp{GNN} from eigendecomposition. However, they are more common in physical systems.

\paragraph{Graph Capsule Networks.}
The idea of Graph Capsule Networks, proposed by \cite{verma2018graph}, is to capture more informative local and global features through spectral convolutions. This is achieved using a capsule vector that contains sufficient discriminative features to enable proper reconstruction. These vectors are constructed using a capsule function that maps the node features to higher-order statistics depending on the given dimensionality of the capsule vector. In a simple form, this could mean that the resulting vector contains the mean and standard deviation of the node's neighborhood. The second key component of graph capsule networks is the aggregation function, which is based on the covariance of a graph and provides information such as norms or angles between node features.
\paragraph{Common Architectures for power grid problems.}
\label{gnn_architectures}
The literature on \ac{GNN} approaches to power grid problems, such as power flow, optimal power flow (OPF), and stability assessment, reveals a set of preferred architectures and design patterns that prioritize accuracy, robustness, and physical fidelity.

The most frequently used architectures include the above-mentioned \ac{GAT}, e.g. \cite{varbella2024powergraph}, \cite{safepowergraph}, and \cite{berezin2024zero}, which often yield the best performance across various grid sizes due to their ability to assign different attention weights to critical nodes. GraphSAGE, as described above, is valued for its scalability and robustness, making it a promising approach for inductive learning on large systems. This is done by e.g. \cite{talebi2025graph}, \cite{safepowergraph} \cite{berezin2024zero}. The above-described spatial graph convolution serves as the fundamental baseline model. For dynamic stability assessment, specialized convolutions, such as ARMA filters, show superior performance. These are based on the auto-regressive moving average and designed to provide a more flexible frequency response \cite{nauck2023toward}.

Successful \ac{GNN} design for power systems is defined by several key choices. In terms of graph representation, most models are bus-centric but often treat the graph as undirected to aid information diffusion. The input features are comprehensive, encompassing the adjacency matrix, nodal features such as injected power and voltage, and edge features including resistance and reactance. For model structure, an Encode-Process-Decode pattern is common. For large grids, models require deep message-passing (up to 48-60 layers) because the physical solution depends on the entire grid topology \cite{piloto2024canos}, although this must be balanced against the oversmoothing challenge, which is described in more detail in Sec.~ \ref{gnn_challgenges}. Finally, there is a strong trend toward Physics-Informed Learning such as \cite{safepowergraph}, \cite{Donon2020} and \cite{jeddi2021physics} described below in Sec.~\ref{gnn_training}.

\subsubsection{Graph Neural Network Training.}
\label{gnn_training}
Since \acp{GNN} are differentiable functions, they can be trained just like ordinary neural networks using gradient descent, backpropagation, batches, or mini-batches. Commonly used loss functions include the negative log-likelihood of softmax functions for the node or graph-level classification. For link prediction, pairwise node embedding losses, such as cross-entropy or Bayesian personalized ranking loss, are common. 
There are several specific considerations to be taken into account when applying \acp{GNN} to power grid problems.
The core distinction in training \acp{GNN} for power grids is the use of Physics-Informed Loss Functions to incorporate domain knowledge. These losses go beyond standard supervised error by actively minimizing the violation of physical constraints like Kirchhoff's Laws and AC-OPF equality/inequality conditions as done in most power grid-related \ac{GNN} studies, e.g. \cite{piloto2024canos}, \cite{varbella2024physics}, \cite{safepowergraph}. This constraint-augmented approach forces the model to learn physically feasible and robust solutions rather than just fitting the training data. Typically, the inequalities or constraints are added to the supervised loss, similar to a regularization term; however, there are also approaches that train solely on the physics loss, as seen in \cite{Donon2020}. The proposed Graph Neural Solver model is trained unsupervised by aiming to directly minimize the violation of Kirchhoff's Laws at every bus in the power grid. The architecture consists of a fixed number of \ac{GNN} layers, each acting as an iterative correction step. These steps push the predicted bus voltage magnitude and angles closer to achieving power equilibrium, thereby satisfying the physical laws.

\subsubsection{Challenges.} \label{gnn_challgenges}
Due to their specific functionality, \acp{GNN} suffer from oversmoothing and oversquashing. Oversmoothing refers to the phenomenon that node features become increasingly similar as the number of layers increases. This problem can be addressed by regularisation or normalization. Oversquashing refers to the distortion of information from distant nodes and is difficult to handle \cite{giovanni2024oversquashing}. In power grid problems, these challenges are primarily addressed through specific architectural choices and computational efficiency. Oversmoothing is addressed by utilizing \ac{GNN} architectures that incorporate skip-connections or ARMA filters \cite{bianchi2021graph}. These connections ensure that node features do not become overly homogenized \cite{ringsquandl_power_2021}, allowing for the deployment of deep models, which are essential because solutions to physical problems depend on the entire grid topology \cite{piloto2024canos}. While this addresses the depth needed for physical fidelity, it is worth noting that literature on certain tasks, such as Neural State Estimation, sometimes finds that shallower models perform better in specific zero-shot scenarios \cite{berezin2024zero}. Regarding oversquashing, the current literature does not explicitly detail specific mechanisms designed to counteract the distortion of information from distant nodes in power grids.
\subsubsection{Scalability}
\label{gnn_scalability}
While general GNNs face significant challenges when processing very large graphs—often requiring complex sampling-based methods or mini-batches of subgraphs, which can lead to exponentially growing computational complexity \cite{ding2022sketchgnn} — this issue is less severe in power grid applications. Physical power systems have a limited and manageable size compared to massive virtual networks, such as social graphs. Even large-scale grids, typically with around 10,000 nodes, can be trained effectively without relying on complicated graph splitting or extensive sampling \cite{piloto2024canos}. Compared to traditional power flow methods, such as Newton-Raphson, which have quadratic or cubic complexity, the scalability of \acp{GNN} stands out as a major strength \cite{hamann2024foundation}. Due to their linear computational complexity, \acp{GNN} achieves a speed-up that can be up to three to four orders of magnitude faster than conventional power flow solvers. For example, the assessment of dynamic stability using \acp{GNN} requires only 1 second, whereas dynamic simulations are up to seven orders of magnitude slower \cite{nauck2023toward}. Furthermore, the use of techniques like GraphSAGE for neighborhood sampling enables efficient training on large grids \cite{talebi2025graph}.

\subsection{Reinforcement Learning}
\label{subsec:rl}
\begin{figure}[htbp]
	\centering
	\includegraphics[width=0.5\linewidth]{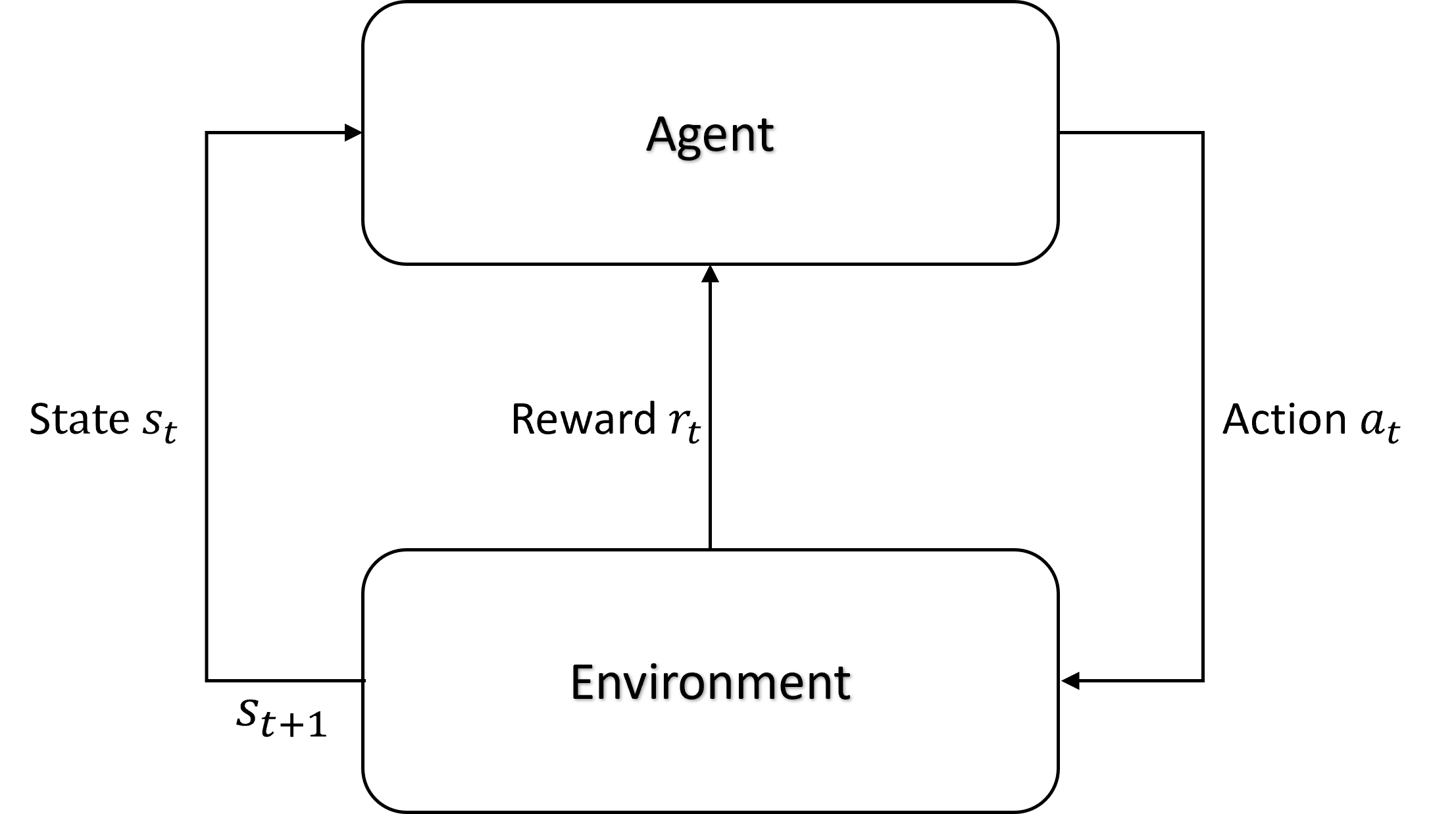}
	\caption{The agent-environment interaction is a cyclical process where the agent selects actions based on the current state, leading to state transitions and rewards, guided by a policy $\pi$, hence generating a sequence of states, actions, and rewards.}
 \label{fig:rl}
\end{figure}

\ac{RL} is a key branch of machine learning that focuses on training agents to make sequential decisions in dynamic environments to maximize cumulative rewards. It primarily uses the framework of \acp{MDP} to model decision-making problems. An MDP is defined by the tuple $M = (S, A, R, T, \gamma, H)$, capturing essential elements of an agent's interaction with its environment. Fig.~ \ref{fig:rl} shows the schematic procedure of a \ac{RL} framework. 

In this formalism, $S$ is the state space of all possible states, and $A$ is the action space of feasible actions. The reward function $R$ maps states and actions to real-valued rewards, providing immediate feedback. The transition function $T$ describes state transitions in response to actions. The discount factor $\gamma$ balances the importance of future rewards against immediate ones. Finally, the horizon $H$ defines the length of an episode, consisting of a sequence of states, actions, and rewards. In contrast, partially observed \acp{MDP} incorporates situations where the agent has incomplete knowledge about the current state. They provide a more realistic framework for many real-world problems where the agent must make decisions based on partial and uncertain information about the environment.  

The primary objective of an agent is to learn a policy function $\pi(s)$ that prescribes actions to states, aiming to maximize the expected cumulative discounted sum of rewards over the time horizon $H$ that defines the length of the episode. Here, $\pi^{*}$ denotes the optimal policy.
\begin{equation}
    	\pi^{*} = \arg\max_{\pi} \mathop{\mathbb{E}}\left[ \sum_{t=0}^{H} \gamma^{t} r(s_t, a_t) \right]
\end{equation}

\ac{RL} algorithms are commonly classified into two main types: model-free and model-based methods. Model-free methods operate without requiring knowledge of the environment's transition functions; instead, they utilize the experiences gathered by the agent. These methods can be subdivided into two primary categories: policy-based and value-based methods, depending on their approach to solving an \ac{MDP}. In contrast, model-based approaches focus on scenarios where the transition function is either known or can be learned. Examples of model-based methods include \ac{MCTS} algorithms like AlphaZero \cite{alphazero}, MuZero \cite{muzero} and EfficientZero \cite{efficientzero}.
In the following, we introduce two important concepts for model-free \ac{RL}, namely value-based and policy-based learning, as well as actor-critic approaches. Then, we will briefly present the widely used model-based technique \ac{MCTS}.

\subsubsection{Model-free Reinforcement Learning}
\label{subsec:modelfree}
\paragraph{Value-based learning.}
Value-based learning estimates the quality of state-action pairs to select optimal actions, i.e., actions with maximum value. For this purpose, the action-value function $Q^{\pi}(s,a)$ represents the expected sum of future discounted rewards, beginning from state $s$, executing action $a$, and subsequently adhering to a given policy $\pi.$ 

\begin{equation}
    	Q^{\pi}(s,a) = \mathop{\mathbb{E}}\left[ \sum_{t=0}^{\infty} \gamma^{t} r(s_t,a_t) \mid \pi, s_0 = s, a_0 = a \right]
\end{equation}

The value function has a key recursive property linking the value of state $s$ to the values of subsequent states $s'$, which is fundamental to many value-based \ac{RL} techniques. This is expressed by the Bellman equation

\begin{equation}
    	Q^{\pi}(s,a) = r(s,a) + \gamma \sum_{s'} p(s,a,s')  \max_{a'}Q^{\pi}(s',a')
\end{equation}
where $p(s,a,s')$ models the state transition dynamics. In value-based approaches, finding the optimal policy involves identifying the optimal value function $Q^{*}(s,a) = \max_{\pi}Q^{\pi}(s,a)$. Explicit solutions to the Bellman equation are possible only when the dynamics function is known \cite{Sutton1999}. Therefore, approximation methods are typically used. Here, we present two such methods.

\textbf{Q-Learning} aims to derive an optimal policy by directly updating values in a Q-table, a lookup table where each entry $Q(s,a)$ estimates the expected cumulative reward for taking action $a$ in state $s$ \cite{watkins1992}. Q-Learning approximates the optimal action-value function $Q^*$ through the following iterative updates.
\begin{equation}
\begin{split}
        	& Q(s,a) \leftarrow Q(s,a) +\\ 
            & \alpha \left[ r(s,a) + \gamma  \max_{a}Q(s',a) - Q(s,a) \right]
\end{split}
\end{equation}
Here, the agent explores the environment with a behavior policy, updating the Q-table based on the discrepancy between the actually observed and the previously expected reward. 
\textbf{\acp{DQN}} uses neural networks to approximate action value functions in high-dimensional input spaces, minimizing the error between current and target Q-values \cite{mnih2015}. It uses two networks, one to select actions and another to compute target Q-values. The target network is periodically updated with the weights of the primary network to stabilize the training. The agent stores its experience in a replay buffer from which it draws samples to train the neural network. Variants such as \ac{DDQN} \cite{ddqn}, dueling \ac{DQN} \cite{duelingdqn}, and Rainbow \cite{rainbow} further improve performance by addressing overestimation and efficiency issues.

\paragraph{Policy-based learning.}
Policy-based learning directly estimates policies without intermediate value functions. It optimizes parameterized policies $\pi_\theta(a \mid s,\theta)$ that specify the probability of action $a$ given state $s$ and parameters $\theta$ to maximize the expected cumulative rewards. A policy can be any mapping from state to action, for example, a neural network. Unlike value-based approaches, policy-based methods update parameters using gradient-based optimization and are suitable for continuous action spaces and stochastic policies \cite{Sutton2018}. 

The objective function $J(\theta)$ aims to maximize the true value function $v_{\pi_\theta}(s_0)$ from the initial state $s_0$. According to the policy gradient theorem \cite{Sutton1999}, $J(\theta)$ is proportional to the sum of the action-value function multiplied by the gradient of the policy:

\begin{equation}
J(\theta) \propto \sum_{s} \mu(s) \sum_{a} q_\pi(s,a) \nabla \pi_\theta(a \mid s,\theta)
\end{equation}

Here, $\mu(s)$ is the distribution under $\pi$, $q_\pi(s,a)$ is the action-value function, and $\nabla \pi_\theta(a \mid s,\theta)$ is the gradient of $\pi$ with respect to $\theta$. The update of the policy parameters proceeds in the direction of the gradient of the objective function to be maximized: $\Delta\theta = \alpha \nabla_\theta J(\theta)$, where $\alpha$ is the learning rate.

\paragraph{Actor-Critic methods.}
These algorithms combine the strengths of value-based and policy gradient-based learning. The actor learns policies to maximize rewards, while the critic evaluates these policies by estimating the value function. This framework addresses the limitations of both approaches and is fundamental to various \ac{RL} algorithms, including \ac{A3C} \cite{A3C}, \ac{DDPG} \cite{DDPG}, \ac{PPO} \cite{PPO}, and \ac{SAC} \cite{SAC}.

\ac{A3C} updates policy and value networks asynchronously through multiple agents, using an advantage function to reinforce better-than-average actions. Further, it applies entropy regularization to enhance exploration, i.e., try new actions rather than exploiting knowledge already gained. Similarly, \ac{DDPG}, tailored for continuous action spaces, simultaneously learns a state-action value function (critic) and a policy (actor), employing experience replay and target networks. \ac{PPO} uses a clipped surrogate objective for smooth policy updates that balance exploration and exploitation, making it a popular choice in RL research. The clipping mechanism limits the policy update to a constrained range, preventing large, potentially destabilizing updates and improving training stability. Finally, \ac{SAC} combines actor-critic methods with entropy regularization, training a policy network and two Q-value networks concurrently to encourage diverse action exploration and reduce overestimation.

While established algorithms such as \ac{PPO} have been used extensively, recent innovations are often overlooked, particularly in power grid control. For instance, \ac{BBF} \cite{bbf} is an advanced method that trains large neural networks in a sample-efficient manner. It employs a ResNet architecture with widened layers, a high replay ratio \cite{Fedus2020} with periodic network resets, and adaptive strategies like dynamic update horizon and discount factor schedules. \ac{BBF} discards NoisyNets \cite{noisynet} in favor of weight decay for regularization, and outperforms state-of-the-art agents in both computational efficiency and performance, enhancing \ac{DRL} for constrained environments.

\subsubsection{Model-based Reinforcement Learning}
\label{subsec:modelbased}
\paragraph{Monte Carlo Tree Search.}
Both value-based and policy-based approaches in \ac{RL} operate as model-free methods, meaning they do not utilize the environment model to plan ahead by simulating future steps. This is where \ac{MCTS} \cite{MCTS} comes into play; it is a heuristic search that combines the accuracy of tree search with the power of random sampling to explore large state spaces efficiently. The algorithm builds a search tree incrementally, where nodes represent states and edges represent actions.

The process begins with selection, where the algorithm chooses the most promising child nodes from the root until it reaches a leaf node. If the leaf node is not terminal, the expansion phase adds one or more child nodes. Next, the algorithm runs a simulation from these new nodes to a terminal state, typically using random actions. Finally, in backpropagation, the simulation results are used to update the values of the nodes in the path from the leaf to the root, propagating the success or failure of the simulation.

\ac{MCTS} effectively balances exploring new actions and exploiting known high-reward actions using the Upper Confidence Bound for Trees (UCT) formula to select nodes. This balance has made MCTS a powerful tool, with notable implementations such as AlphaZero \cite{alphazero}, MuZero \cite{muzero}, and EfficientZero \cite{efficientzero} achieving superhuman performance in complex games.

Designed to master complex games such as chess, shogi, and go, AlphaZero uses deep neural networks combined with \ac{MCTS} and relies on predefined game rules. It learns by playing and using \ac{RL} \cite{alphazero}. MuZero extends this approach by generalizing to environments with unknown rules, integrating \ac{RL}, \ac{MCTS}, and learned models to predict the environment dynamics \cite{muzero}. EfficientZero builds on MuZero and achieves superhuman performance on the 100k Atari benchmark, significantly outperforming previous state-of-the-art results \cite{efficientzero}. It introduces innovations such as self-supervised consistency losses for accurate next-state prediction and end-to-end value prefix prediction to deal with state aliasing issues. These enhancements improve exploration and action search capabilities, making EfficientZero highly effective in data-limited scenarios.

\section{Graph Reinforcement Learning for Transmission Grids}
\label{chap:TG}
To ensure safe and reliable transmission, human experts manually manage power grids. However, the rise in renewable energy and demand necessitates automated, data-driven optimization \cite{marot2021learning}, shifting grid operation to adapt to generated power rather than predicted demand \cite{viebahn2022potential}.

Many transmission grid control methods focus on generation or loads, like re-dispatch \cite{Kamel2020,Bai2023,enlite} or load shedding \cite{Larik2019loadsheddingreview}. Topology actions, such as substation reconfiguration through busbar splits \cite{silver2017mastering}, offer a cost-effective alternative and enable an efficient power flow rerouting. In this way, curtailment of renewable energy can be avoided to some extent, and more power can be transmitted using the same infrastructure. 

Solving this topology control problem is a significant challenge. It is an inherently non-linear, non-convex, and large-scale combinatorial problem. Classical methods, such as linear programming, quadratic programming, and non-linear programming, often rely on convex relaxation, linearization, or assumptions of continuity in the objective function. Yet they face critical limitations that render them unsuitable for modern transmission grid operations. First, classical optimization methods struggle with the non-convexity and non-linearity of AC power flow equations, which are essential for accurate grid modeling in close-to-real-time operations \cite{Lee2019convexrestriction}. Secondly, mathematical solvers like IPOPT or CPLEX may fail to converge within the tight time windows required for real-time operations, such as during cascading failures or sudden renewable generation drops \cite{CAPITANESCU2013276, mohagheghi2018}. For many critical grid operations, such as topology control, these traditional solvers are computationally intractable \cite{rajaei2025transferable}.

The core advantage of \ac{GRL} over these high-performance mathematical solvers (e.g., MISOCP or MIP) is not necessarily in finding a provably optimal solution, but in its ability to find high-quality, feasible solutions within a near-real-time decision window. For practitioners, this level of performance is often more than sufficient, as timely, robust decisions typically outweigh the marginal gains of exact optimality in transmission grid operation. A GNN-accelerated approach can deliver feasible solutions orders of magnitude faster than mathematical solvers \cite{rajaei2025transferable}. It further allows for adapting to uncertainty in real-time in a way that is impossible for solvers that must recompute from scratch for every new scenario \cite{marot2021learning, rajaei2025transferable}. This massive gain in speed, scalability, and adaptability is the primary driver for \ac{GRL}.

\ac{DRL}-based grid operation approaches are validated through simulations, often using the Grid2op \cite{grid2op} environment from the Learning to Run a Power Network (L2RPN) challenges \cite{L2RPN}. Grid2Op is crucial for developing methods that tackle grid congestion and enhance reliability; however, it is an abstraction of reality and may potentially limit its real-world applicability. Within this context, \ac{GRL} has emerged as a promising paradigm that leverages the structural properties of power grids. The following section reviews existing \ac{GRL} approaches for transmission grid control, highlighting their algorithmic design, graph representations, and evaluation setups. Furthermore, we discuss potential pitfalls that arise from overreliance on Grid2op and outline mitigation strategies at the end of the chapter.

We identified eleven \ac{GRL} approaches for managing transmission grids in real-time. Due to a robust pre-dispatch schedule also given in real-world grid control, actions are necessary only in critical states. In stable conditions, agents usually do not act and only intervene when the line loading exceeds a threshold. This procedure is common across many machine learning approaches \cite{binbinchen,lehna2024hugo,curriculumagent,taha_learning_2022,yoon2021winning, sar_multi_agent_2023,dejong2024, dejong2025, hassouna2025_soft}. Tab.~\ref{table:transmission_grid} lists the \ac{RL} method, action type, GNN architecture, grid size, and overall focus of the analyzed approaches, and Fig.~\ref{fig:grloverview} illustrates the logical flow in \ac{GRL} for grid control. A clear research trend is the gradual shift from monolithic to hierarchical and multi-agent designs, as well as the use of attention-based GNNs to improve interpretability and generalization.

\newcolumntype{P}[1]{>{\raggedright\arraybackslash}p{#1}}
\newcolumntype{M}[1]{>{\centering\arraybackslash}m{#1}}
\begin{table*}
\centering
\renewcommand*{\arraystretch}{1.4}

\caption{\textbf{Overview of \ac{GRL} approaches proposed for transmission grids.} We categorize approaches by the used \ac{RL} algorithm and \acp{GNN}, specifying action types and emphasizing topological actions (Topo) alongside others. Grid sizes indicate the number of buses evaluated. The focus column describes distinctive aspects of each approach.}
\label{table:transmission_grid}
\begin{tabular}{|M{0.8cm}||P{1.9cm}|P{2.1cm}|P{2.2cm}|P{2.2cm}|P{1.5cm}|P{2.7cm}|P{4.8cm}|}

\hline
env
& {Approach} 
& {Control Algorithm} 
& {Action} 
& {GNN}
& {Grid Size}
& {Focus}
\\
\hline
\hline
 \multirow{6}{*}{{\begin{turn}{90}grid2op \,\,\,\,\,\,\,\,\,\,\,\,\,\,\,\,\,\,\,\,\,\,\,\,\,\,\,\,\,\,\,\,\,\,\,\,\,\,\,\,\,\,\,\,\,\,\,\,\,\,\,\,\,\,\,\,\,\,\,\,\,\,\,\,\,\,\,\,\,\,\,\,\,\,\,\,\,\,\,\,\end{turn}}}   & Xu et al. 2020 \cite{xu2020}&Double-Q-Network&Topo&GCN& 14&Simulation-constrained RL\\
\cline{2-7}

                                                        &Taha et al. 2022 \cite{taha_learning_2022} &MCTS for action selection&Topo&GCN with residual connections&  118 (2020)&GCNs for power flow estimation\\
\cline{2-7}

                                                        &Sar et al. 2023 \cite{sar_multi_agent_2023}& Multi-Agent SACD\& PPO &Topo&GCN&  5&Hierarchical RL\\
\cline{2-7}

                                                        &Yoon et al. 2021 \cite{yoon2021winning}&SMAAC&Topo&Transformer GNN&  5, 14, 118 (2020)&Afterstate representation, goal topology actions \\
\cline{2-7}

                                                        &Qui et al. 2022 \cite{qiu_distribution_2022}&SMAAC&Topo&GAT&  5,14, 118 (2020 subset of 36 substations)& Afterstate representation, attention mechanism \\
\cline{2-7}
                                                        &Fabrizio et al. 2025 \cite{fabrizio2025}&Distributed Dueling DQN + DQfD&Topo&Shared GAT (line graph)& 14 & Distributed RL, pre-training, potential-based reward shaping\\
\cline{2-7}
                                                        &Anguiano-Batanero et al. 2025 \cite{anguiano2025graph}&PPO&Topo&Transformer GNN& 5 & Action Masking, multiple graph representations\\
\cline{2-7}
                                                        &Peter et al. 2025\cite{peter2025dualpolicy}&PPO &Topo&GCN& 14 & Dual-policy, N-k contingency\\
\cline{2-7}

                                                        &Xu et. al 2022 \cite{xu_active_2022}&Double-Dueling DQN&Topo \& redispatch&GAT&  118 (2020)& MCTS for action space reduction, sub action spaces for multiple agents\\
\cline{2-7}
                                                        &Zhao et al . 2022 \cite{zhao2022graph}&PPO&Redispatch, curtailment,  battery storage&GraphSAGE&  118& Representation of Power Grids, simulation of bus additions \\
\cline{1-7}

\multirow{2}{*}{{\begin{turn}{90}other\,\,\,\,\,\,\,\,\,\,\,\end{turn}}}      &Wu et al. 2023 \cite{wu_constrained_2023}&Primal-Dual Constrained TD3&Active \& reactive power control, battery operation&Cplx-STGCN&  14, 30 & Feasible control for SDOPF optimization \\
\hline

\end{tabular}
\vspace*{2mm}

\end{table*}

\subsection{RL Framework}
All \ac{GRL} approaches reviewed in this chapter formulate the transmission grid control task as a Markov Decision Process (MDP), as introduced in Sec. \ref{subsec:rl}. The following sections detail how these approaches specifically define the core components of this framework: the states ($S$), actions ($A$), and rewards ($R$).
\paragraph{Rewards.}
The overall goal of all approaches analyzed here is to manage line flows while mitigating congestion and minimizing overall costs. The reward functions, which guide the \ac{GRL} agents, can be categorized into three objectives, as detailed in Table \ref{tab:transmission_metrics}. 

\textbf{Firstly, Grid Stability and Survival:} Several approaches prioritize keeping the grid operational under high congestion. For instance, \cite{xu2020,xu_active_2022} and \cite{taha_learning_2022} explicitly penalize line overflows or heavy loading ($\rho > 0.9$) to prevent cascading failures. Similarly, \cite{fabrizio2025} and \cite{peter2025dualpolicy} design custom survival rewards to encourage agents to maintain grid stability for as long as possible during extreme events. Specifically, \cite{peter2025dualpolicy} comprise three components: a logarithmically scaled survival reward, an overload penalty proportional to the number of over-threshold lines, and an action term that favors conservative do-nothing policies unless line loadings exceed a critical threshold. This formulation encourages stability through minimal interventions. \cite{fabrizio2025} augment their survival reward with potential-based reward shaping \cite{adamczyk2025} to improve credit assignment across distributed agents.
\textbf{Secondly, Grid Efficiency:} A second cluster of works focuses on maximizing the composite L2RPN Score. Since this score heavily penalizes blackouts, these approaches inherently prioritize survival as a prerequisite for optimizing efficiency. Approaches such as \cite{yoon2021winning,sar_multi_agent_2023} and \cite{qiu_distribution_2022} reward the ratio of generated to served electricity, incentivizing the agent to reduce transmission losses while ensuring the grid remains operational. \cite{anguiano2025graph} also targets efficiency by minimizing power loss through topological changes.
\textbf{Thirdly, Operational Constraints:} Finally, approaches dealing with continuous control variables, such as \cite{zhao2022graph} and \cite{wu_constrained_2023}, utilize complex reward functions that combine operational costs (e.g., generator dispatch) with soft penalties for physical constraint violations, such as voltage limits or power flow equations.

Overall, most GRL agents use composite reward formulations balancing congestion penalties with efficiency and action costs. However, no standardized reward definition exists, making cross-comparison difficult.
\paragraph{Actions.}
In critical states, agents can re-dispatch or alter the grid topology by changing bus configurations or line connectivity, often reconnecting disconnected lines. The actions considered by each approach are listed in Tab.~\ref{table:transmission_grid}. Since topology actions can often be done by the grid operators themselves and, thus, are cheaper than redispatch actions, they are favored if possible. However, the inherent combinatorial complexity of real-world busbar configurations makes it impractical to simulate every configuration in larger grids. Grid2Op curtails this by excluding electrically symmetrical actions, as well as any actions that would lead to a disconnected or islanded grid. While this reduces the action space, it remains large nonetheless. All works limit the enormous topological action space via masking, reduction, or hierarchical control. None has yet attempted to learn action embeddings or continuous latent actions.

\paragraph{States.}
Most approaches use the information provided by Grid2op, although typically not all features are used. For most approaches, the state includes grid topology, connected elements, and features such as bus-bar data, generation and loads, voltages, and line flows. Furthermore, the ratio ($\rho$) between the current flow and the thermal limit of each line is a critical feature. Typically, the states are modeled as graphs embedded using a \ac{GNN} (see Sec.~\ref{subsec:TG_GNN}). The state information is consistent primarily across the \ac{GRL} approaches, only \cite{wu_constrained_2023} operate in a different environment and use only voltages as states. 
 
\subsection{Overall Approach and RL Algorithms}
\label{subsec:TG_approaches}
While the analyzed \ac{GRL} frameworks share similar state, action, and reward structures, they diverge significantly in their core \ac{RL} algorithm and architectural design. These approaches can be broadly categorized by how they structure the decision-making process and how they contend with the massive combinatorial action space and safety constraints inherent in grid control. We can identify several key trends: a move from monolithic agents to hierarchical or multi-agent architectures, the use of planning algorithms to guide exploration, and the integration of imitation learning to accelerate training.

\paragraph{Monolithic Agents and Action Space Management}
A first group of approaches uses a single (monolithic) agent but employs specific techniques to make the decision-making process tractable and safe.

Early work by \cite{xu2020} employ value-based methods (see Sec.~\ref{subsec:modelfree}) with safety layers. Recognizing that naive exploration violates constraints, they introduced a "soft constraint" to replace invalid actions with a "do-nothing" action. This allows exploration without triggering constraint violations. During exploitation, they verify the top N actions with the highest Q-values of their Double-Q-learning agent through power flow simulation, creating a safety layer that prevents catastrophic failures.

A more direct method for ensuring safety is dynamic action masking. \cite{anguiano2025graph} implement a purely model-free \ac{PPO} agent (cf. Sec.~\ref{subsec:modelfree}) equipped with a "Topological Action Converter" (TAC). This module dynamically masks all infeasible or redundant topological actions at the policy output.  This action-masking mechanism, implemented within \ac{PPO}, ensures that only valid busbar configurations are selected during training and inference. This effectively reduces the exploration space without sacrificing flexibility.

Tackling the problem from a formal optimization perspective, \cite{wu_constrained_2023} address the stochastic dynamic \ac{OPF} problem with \ac{RES} and decentralized energy systems.  They use an actor-critic method where separate neural networks predict voltages. The critic networks are refined with temporal difference learning. They integrate constraints with Lagrangian multipliers, leveraging the duality principle to optimize both primal and dual variables through gradient-based updates. This approach is also used in similar constrained \ac{RL} problems, such as in \cite{yan_multi_2023}.

\paragraph{Planning-Based and Model-Based Strategies}
Instead of learning a policy directly from rewards, a second group of methods adopts the model-based \ac{RL} paradigm from Sec. \ref{subsec:modelbased}. They use planning algorithms like Monte Carlo Tree Search (MCTS), a model-based heuristic search technique to explore the action space more intelligently. 

Instead of Grid2op simulations, \cite{taha_learning_2022} took a model-based \ac{RL} direction. They first train a \ac{GNN} to predict the resulting grid state, i.e., line loading $\rho$ for different topologies. They estimate the state evolution under different actions and select the trajectory maximizing cumulative rewards using \ac{MCTS}. They iteratively build a tree, starting with the initial state as the root and actions as edges leading to subsequent nodes. This directly applies the MCTS framework described in Sec. \ref{subsec:modelbased}, where the tree search effectively prunes the vast combinatorial action space by simulating future states rather than relying solely on a learned policy. Only nodes with sufficiently low loads are retained, and a given number of steps is simulated using a 'do-nothing' agent to determine the node's value. The \ac{GNN} predicts line loading for each action. After simulating steps with a do-nothing agent, they select optimal actions by maximizing node values and the number of possible future actions.

\cite{xu_active_2022} present another \ac{MCTS}-based approach with a similar tree structure. Rather than a learned model, they use the simulator to build the tree. The leaf nodes represent overload states from feasible actions; the highest-value path determines the best action. Using double-dueling Q-networks, they train multiple sub-agents to select actions from different sub-action spaces obtained by dividing the reduced action space from \ac{MCTS} into a fixed number of subspaces.
The MCTS search effectively reduces the vast action space, creating a hybrid between a monolithic planner and a multi-agent execution system. A long-short-term strategy balances immediate and long-term benefits and manages sub-agents effectively. Each agent simulates $n$ actions at overload states, selecting the best through efficient comparison. They also constrain topological actions and re-dispatch to stay within physically plausible limits.

\paragraph{Hierarchical and Multi-Agent Decomposition}
To manage the sheer combinatorial complexity of the grid's action space, a prominent trend is to decompose the monolithic \ac{MDP} (Sec. \ref{subsec:rl}) into smaller, manageable sub-problems. This is achieved through hierarchical and multi-agent architectures, which are purpose-built to divide the control task. This decomposition can be based on the \textbf{type of decision} or the \textbf{grid's physical components}.


Basing the decomposition on the the type of topological action, \cite{yoon2021winning} and \cite{qiu_distribution_2022} propose a hierarchical policy. A high-level agent generates a desired goal topology rather than a specific action. A low-level policy is then responsible for executing the changes to reach that goal. This strategy avoids learning individual actions by focusing on suitable topologies for critical situations. Both utilize an afterstate representation to capture the grid topology after a topological action, which is advantageous when sequences of actions lead to identical topology changes. This directs their \ac{RL} algorithm to understand the stochastic dynamics following each action. They use rule-based approaches like CAPA to prioritize substations with high-capacity usage and ensure timely responses. Their actor-critic algorithm enhances exploration and value function determination using the afterstate representation and goal topology predictions. \cite{qiu_distribution_2022} utilizes the whole architecture and present a similar approach with a different attention mechanism in their \ac{GCN}.

On the other hand, \cite{peter2025dualpolicy} decompose the problem based on the operational scenario. They train two \ac{PPO} policies: a "general" policy for stable conditions and a "critical" policy for emergency states, which is activated when line loading exceeds a threshold. This separation allows each policy to specialize in different operational regimes, improving robustness against extreme contingencies. Both policies share a \ac{GCN} state encoder but have separate actor-critic networks, allowing each to specialize. To further stress-test the agent, an opponent model randomly disconnects multiple lines to emulate cyberattacks or cascading failures, effectively simulating $N$–$k$ events up to $k=5$. The agent learns to mitigate these disruptions through topological switching actions while minimizing unnecessary interventions during normal operation. This dual-policy setup conceptually resembles hierarchical RL structures but differs in that it maintains fully independent actor–critic networks for each regime rather than nested decision levels. This architecture does not necessarily address the action space exploration problems if applied to a larger grid.

Decomposing the problem based on grid components, \cite{sar_multi_agent_2023} present a three-level hierarchy. A top-level agent decides on the need for action and identifies critically loaded lines.  It activates the mid-level agent in unsafe scenarios that prioritizes high-load substations using CAPA \cite{yoon2021winning}. At the lowest level, substation-specific agents select bus assignments from a predefined action space. This approach is noteworthy for comparing parameter-shared versus independent critics in \ac{PPO} and \ac{SAC}.

Similarily, \cite{fabrizio2025} introduce a fully distributed framework. A high-level manager coordinates a team of low-level agents, each responsible for controlling a single power line. This deep decomposition relies on a shared \ac{GNN} to provide local context to each agent. Each low-level agent controls a single power line. Unlike prior substation-based approaches \cite{yoon2021winning,sar_multi_agent_2023,van2023multi}, their model decomposes both the action and observation spaces. Each agent receives a local view processed through a shared \ac{GNN}, which encodes neighborhood information and mitigates partial observability. The high-level controller—also a Dueling DQN—decides which line-level agent acts at each step.

\paragraph{Hybrid Approaches with Imitation Learning}
Finally, several approaches accelerate the difficult learning process by leveraging expert data through imitation learning. 

\cite{fabrizio2025} pre-train their distributed agents using Deep Q-learning from Demonstrations (DQfD) on expert trajectories and further refine the policies through bootstrapped reward shaping. This allows the agents to learn a competent policy from expert trajectories before exploring on their own. This setup promises a scalable, modular coordination and enables GNN-based transfer learning across grid sizes.

To address the variability in grid topologies caused by, e.g., extreme weather or maintenance, \cite{zhao2022graph} focus on re-dispatching, curtailment, and battery storage to ensure stability. Similar to \cite{binbinchen,lehna2024hugo}, they use imitation learning to pre-train a \ac{PPO} agent with a \ac{GNN} based on GraphSAGE \cite{GraphSAGE}.   

In summary, the field is clearly evolving from applying standard, monolithic \ac{RL} algorithms toward designing sophisticated, structured architectures. These hierarchical, multi-agent, and planning-based systems are purpose-built to decompose the high-dimensional, combinatorial, and safety-critical nature of power grid control. Furthermore, pre-training agents using imitation learning has provided very promising results in pure imitation learning studies \cite{dejong2025,hassouna2025_soft}, refining policy initialization and improving sample efficiency in subsequent \ac{RL} phases. Consequently, the combination of imitation learning and \ac{RL} is increasingly explored to leverage the strengths of both paradigms—using expert demonstrations to guide early learning while allowing agents to further optimize their performance through exploration and interaction with the environment. 

\subsection{Graph Embeddings}
\label{subsec:TG_GNN}

In all analyzed approaches, the power grid state is modeled as a graph, and a \ac{GNN} is used to learn a latent representation that captures both the grid's features and its current topology. While this general procedure is universal, the approaches differ significantly in three key areas: the graph representation, which defines what is modeled as nodes and edges; the GNN architecture, which determines how information is processed within the graph; and the role of the GNN, which specifies the purpose of the learned embedding, such as for policy learning, value estimation, or world modeling.

\begin{figure*}[htbp]
	\centering
	\includegraphics[width=.9\textwidth]{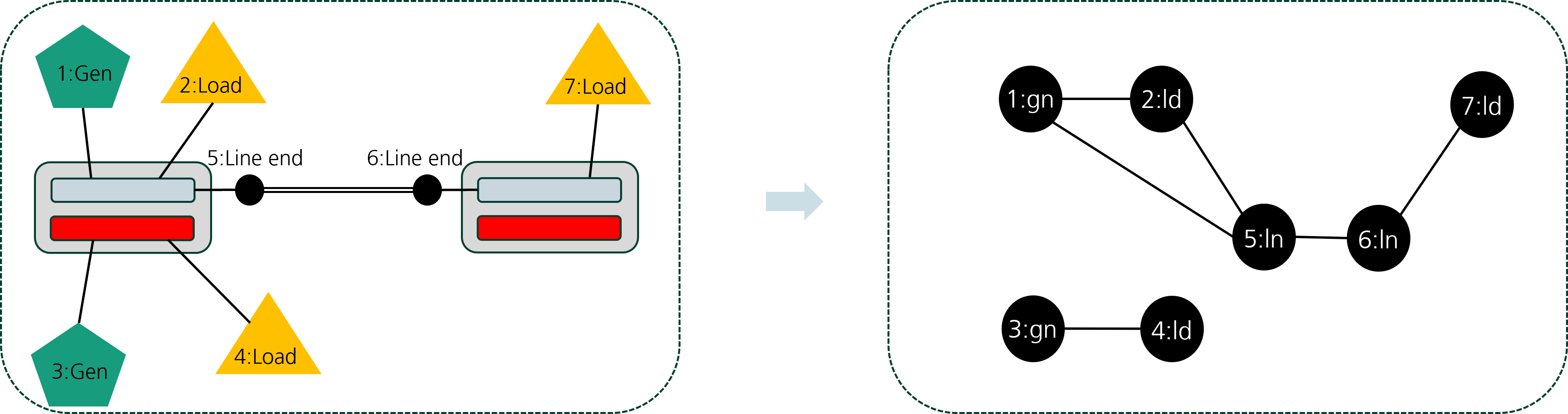}
	\caption{Transformation from the physical power grid to the graph input for the GNN. Each grid component — loads, generators, and both ends of transmission lines — is represented as a node. Edges are defined by the grid’s physical connectivity, linking nodes within substations and across substations via transmission lines.}
 \label{fig:graph_representation}
\end{figure*}

\paragraph{Defining the Graph Representation}
The choice of graph structure is critical, as it defines the information available to the \ac{GNN}. Most approaches use the graph representation illustrated in Fig.~\ref{fig:graph_representation}. It is the default graph structure provided by \texttt{Grid2Op} \cite{sar_multi_agent_2023,xu2020,peter2025dualpolicy}, where nodes represent loads, generators, and the two ends of transmission lines. These nodes are connected according to their busbar assignments within substations, and the endpoints of each transmission line are also connected. This design naturally captures possible connections across busbars and facilitates the representation of substations that may be split. In some variants, transmission lines are instead modeled as edges equipped with line-specific features, while nodes represent only the electrical buses. Node features in both cases typically include all available state information. Fig.~\ref{fig:graph_representation} illustrates how a topological configuration of two substations is transformed into a graph representation.

Moreover, \cite{anguiano2025graph} provide a systematic comparison of representations, evaluating three levels of abstraction: a flat representation, which uses a raw observation vector without any graph structure; a \textit{SubstationGraph}, which models substations as nodes and transmission lines as edges; and an \textit{ElementGraph}, a detailed representation similar to the \texttt{Grid2Op} default, where each component—load, generator, and line —is represented as a node. Their findings show that the most detailed representation, the \textit{ElementGraph}, achieves the best performance and stability, underscoring the importance of incorporating fine-grained structural information into the \ac{GNN}.

A key challenge in grid modeling is the busbar information asymmetry \cite{dejong2025}, where the default graph structure (cf. Fig.~\ref{fig:graph_representation}) only passes messages between elements connected to the same busbar, while messages from elements on the other busbar within the same substation are not passed. This limits the model's expressiveness, as it makes elements on a different busbar (but at the same substation) indistinguishable from objects at entirely different substations. To address this, \cite{fabrizio2025} propose a homogeneous line-graph. In this elegant representation, each power line is represented as a node, and edges connect lines that share a common substation. This design inherently addresses the asymmetry problem and facilitates a distributed agent framework where each agent controls one line.

\paragraph{Graph Neural Network Architectures}
Given a graph representation, the next choice is the \ac{GNN} architecture used to extract features.
A common baseline is the \ac{GCN}, a spectral \ac{GNN} defined in \ref{equ:spec_approx}. Basic 2 or 3-layer \acp{GNN} are used by \cite{sar_multi_agent_2023,xu2020,peter2025dualpolicy} as a robust encoder. \cite{taha_learning_2022} also use a GCN, but augment it with residual connections to create a deeper 3-layer model for predicting system dynamics.

Furthermore, a clear trend is the adoption of attention mechanisms to allow the model to weigh the importance of different neighbors. \cite{xu_active_2022} directly compare \ac{GAT} models against \ac{GCN} and find that the \ac{GAT} architecture leads to longer stable operation. Leveraging the attention mechanism defined in Eq. \ref{equ:att_conv} (Sec. \ref{subsubsec:messagepassing}), this architecture allows the agent to dynamically weigh the importance of connected grid elements—crucial for identifying specific lines that are nearing thermal limits—rather than weighting all neighbors equally as in standard spatial convolutions. The model in \cite{fabrizio2025} is also \ac{GAT}-based, using attention to enrich each line agent's local observation with contextual information from neighboring lines. \cite{yoon2021winning} and \cite{anguiano2025graph} employ transformer attention, which is an improved variant of attention convolutions (cf. Eq. \ref{equ:att_conv}). It introduces nonlinearity to the computation of the attention coefficients, allowing for more diverse coefficients. Similar to the original transformer \cite{vaswani2017attention}, it includes additional weight matrices that project the features from the node itself and neighboring features separately. This enhances the handling of edge features and enables deeper GNN architectures due to improved convergence properties. However, it also increases the computational complexity. 

Some works employ architectures designed for specific challenges, such as GraphSAGE and spatio-temporal models.\cite{zhao2022graph} use GraphSAGE \cite{GraphSAGE}, an inductive \ac{GNN} that learns by sampling and aggregating features from a node's local neighborhood as described in \ref{graph_sage}. By utilizing the neighborhood sampling technique, the model avoids processing the entire graph simultaneously. This directly addresses the scalability and overfitting challenges discussed in Sec. \ref{gnn_challgenges}, allowing the agent to handle larger grid variations without retraining. This is explicitly chosen to improve generalization to unseen grid topologies, as GraphSAGE has been applied in \ac{GNN}-based imitation learning approaches for transmission grids \cite{dejong2025}. \cite{wu_constrained_2023} introduce the only spatio-temporal model, a \ac{STGCN} trained using a physics loss that represents the violation of the power flow equation as described in Sec.~\ref{gnn_training}. It models the dynamics in power grid topologies by combining temporal convolutional layers with spatial convolutional layers to capture both types of dependencies. The temporal 1D-CNN layer extracts time-dependent features, and the spatial graph convolutions use the grid's admittance matrix as a graph shift operator to learn node embeddings. In both types of layers, the inputs are complex-valued, and the imaginary and real components are processed separately.

Overall, anywhere high robustness and interpretability are critical, attention-based \acp{GNN} are the prevailing choice. Benchmarks show GAT as superior across various scenarios, effectively handling disturbances. Conversely, GraphSAGE is the preferred architecture when the main goal is generalization to unseen topologies and scalability.

\paragraph{Role of the Graph Neural Network in the RL Framework}

Finally, the learned embeddings are used in different ways within the \ac{RL} loop, reflecting distinct algorithmic philosophies. 

The most common approach is to use the \ac{GNN} as a \textbf{shared policy/value encoder}. In this setup, the \ac{GNN} functions as the primary state encoder, processing the raw graph-structured state into a latent embedding vector. This vector is then fed as input to both the actor (policy) and critic (value) networks. This design is implemented with various \ac{GNN} architectures. For instance, \cite{peter2025dualpolicy} utilize the spectral \ac{GCN} output embedding as the common input for their two separate \ac{PPO} algorithm instances. Similarly, \cite{xu_active_2022} use a two-layer \ac{GAT} architecture for both the actor and critic in their \ac{PPO} agent. \cite{anguiano2025graph} aggregates their GraphTransformer node embeddings into a global latent vector, which in turn serves as the input to their \ac{PPO} actor and critic networks.

Sharing the GNN parameters across networks is a common strategy to promote efficient representation learning. \cite{yoon2021winning} shares the learned embeddings across the actor and critic heads. The actor learns the parameters of a normal distribution for action sampling, while the critic transforms the embeddings into a scalar representing the value function of their afterstate representation. This sharing principle is also central in multi-agent settings. \cite{sar_multi_agent_2023} apply shared \ac{GNN} blocks  in both their single and multi-agent frameworks. Likewise, the distributed framework by \cite{fabrizio2025} relies on a shared \ac{GNN} as a common feature extractor for all its line-level agents. This shared embedding enriches each agent's local observation with contextual information from its neighbors, enabling implicit coordination.

In contrast to these end-to-end trained encoders, some approaches decouple representation learning from the \ac{RL} objective. \cite{zhao2022graph} first train their GraphSAGE network in an \textbf{unsupervised manner} to learn structurally meaningful embeddings independent of the \ac{PPO} loss. This pre-trained, "frozen" encoder is then used by the \ac{RL} agent, a strategy chosen to improve generalization across different grid topologies.

A third, distinct role for the \ac{GNN} is as a \textbf{world model for planning}. In a model-based \ac{RL} approach, \cite{taha_learning_2022} use their \ac{GCN} not as a policy encoder, but as a learned physics model. The \ac{GCN} is trained to provide node level predictions of the line loads ($\rho$) for different actions, and this predictive model is then used by an \ac{MCTS} planner to select the best action sequence. For training, they utilize grid topologies that are similar to a reference topology. If the \ac{GNN} shows increased generalization error, they revert to the reference topology, which helps maintain grid stability, as supported by \cite{lehna2024hugo,curriculumagent,hassouna2025_soft}.

In summary, the field is moving beyond using standard \ac{GCN} architectures on default grid graphs. A clear trend is emerging toward more sophisticated, problem-specific representations and more powerful attention-based architectures. Furthermore, the role of the \ac{GNN} is diversifying, being used not just as a policy input, but also as a predictive world model or an unsupervised feature extractor to enhance generalization.

\begin{figure*}[htbp!]
	\centering
	\includegraphics[width=.9\textwidth]{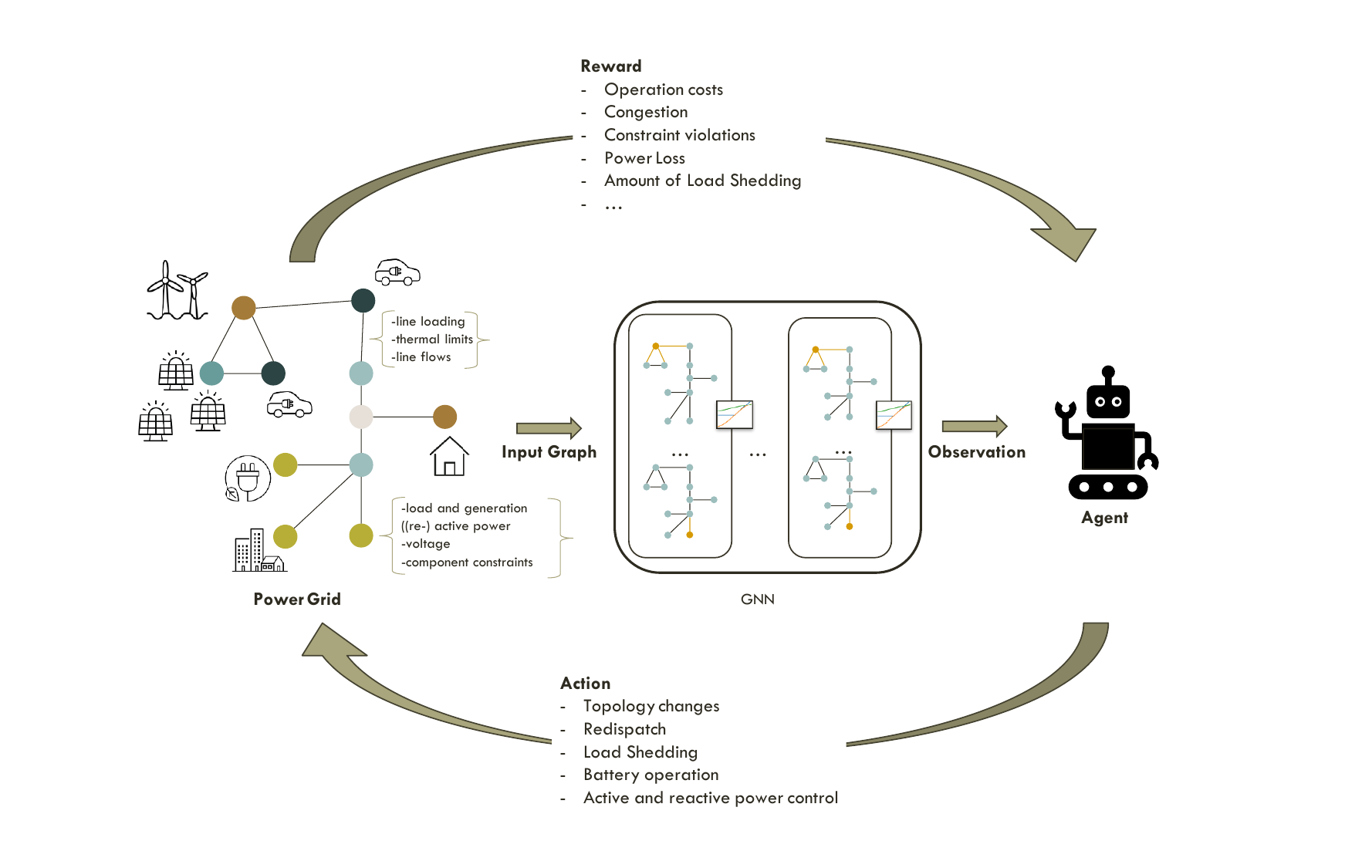}
	\caption{\textbf{Illustration of the Logical Flow of GRL for Grid Operation}: First, the power grid, including relevant information about lines and grid nodes, is modeled as a graph with node and edge features. This graph is then input into a GNN model, which learns a representation of the grid. This representation serves as an observation for the agent. Based on this observation, the agent selects an action from the action space. This may include simulations or other verification strategies to validate the action. The final action is executed in the environment (i.e., the simulated power grid). The agent receives a reward corresponding to the quality of the action and updates its weights. Depending on the RL algorithms employed, multiple (Graph) Neural Networks must be updated; for example, in actor-critic approaches.}
 \label{fig:grloverview}
\end{figure*}
\subsection{Experiments and Evaluation}
Most approaches train and evaluate their agents on grid2op because of its extensive power grid simulations with realistic data. \cite{xu2020,fabrizio2025} focuses on an IEEE 14-bus system with 20 transmission lines, 6 generators, and 11 loads across 1004 scenarios over 4 weeks at 5-minute intervals. \cite{taha_learning_2022}, \cite{xu_active_2022}, and \cite{zhao2022graph} use the larger IEEE 118 grid, while \cite{sar_multi_agent_2023} uses the IEEE 5 grid for a hierarchical multi-agent proof of concept. \cite{yoon2021winning} evaluate on all three of them and\cite{wu_constrained_2023} applies their method to IEEE 14 and IEEE 30 bus systems with wind power data for power flow control using \ac{BESS}. In terms of evaluation, however, the presented approaches on grid control are not as comparable as one could hope, considering that most are based on the same framework and utilize similar data. Table \ref{tab:transmission_metrics} provides a comprehensive summary of the experimental variables used across these studies, clustering approaches by their primary reward objectives (Stability, Efficiency, or Operational Constraints) and detailing the key performance metrics reported.

\begin{table*}[t]
\centering
\small 
\caption{Overview of Reward Variables and Experimental Metrics (Transmission Grid)}
\label{tab:transmission_metrics}
\renewcommand{\arraystretch}{1.2} 
\begin{tabular}{|M{4cm}||P{5.5cm}|P{5.5cm}|}

\hline
\textbf{Reference}
& \textbf{Reward Variables (Optimization Objective)}
& \textbf{Experimental Performance Metrics}
\\
\hline
\hline
\multicolumn{3}{|c|}{\textbf{Grid Stability \& Survival} (\textit{Focus: Maximizing survival time})} \\
\hline
Xu et al. 2020 \cite{xu2020} & Line overflow proximity, Survival bonus (+1/-1), Redispatch cost & Avg. Survival Steps, Successful episodes rate \\
\hline
Xu et al. 2022 \cite{xu_active_2022} & Penalty for Overload ($\rho > 1$) and Heavy load ($\rho > 0.9$) & Avg. Survival Steps, Decision time  \\
\hline
Fabrizio et al. 2025 \cite{fabrizio2025} & Survival reward (+1), Potential-based reward shaping & Survival Time, Inference speed  \\
\hline
Peter et al. 2025 \cite{peter2025dualpolicy} & Logarithmic survival ($ \alpha \log t$), Overload penalty & Survival rate under \textbf{N-k contingencies}  \\
\hline
Taha et al. 2022 \cite{taha_learning_2022} & Binary safety reward ($\rho < 1$) cumulative over MCTS horizon & \textbf{Failure rate reduction}  \\
\hline
\hline

\multicolumn{3}{|c|}{\textbf{Grid Efficiency (L2RPN)} (\textit{Focus: Minimizing energy losses})} \\
\hline
Yoon et al. 2021 \cite{yoon2021winning} & Grid Efficiency ($ \sum Load / \sum Gen$) & \textbf{L2RPN Score}, Survival rate \\
\hline
Sar et al. 2023 \cite{sar_multi_agent_2023} & Grid Efficiency ($ \sum Load / \sum Gen$) & Training score \textbf{convergence} (Multi-agent vs. Single-agent) \\
\hline
Qiu et al. 2022 \cite{qiu_distribution_2022} & Grid Efficiency ($ \sum Load / \sum Gen$) & \textbf{L2RPN Score}, Management steps \\
\hline
Anguiano et al. 2025 \cite{anguiano2025graph} & Power Loss Reduction  & \textbf{Steps to Complete (S2C)} (Convergence metric), Cost savings \\
\hline
\hline

\multicolumn{3}{|c|}{\textbf{Operational Constraints} (\textit{Focus: Voltage limits and dispatch costs})} \\
\hline
Zhao et al. 2022 \cite{zhao2022graph} & Operational costs + Penalties for voltage/flow violations & Reward convergence, \textbf{Topology generalization} (t-SNE) \\
\hline
Wu et al. 2023 \cite{wu_constrained_2023} & Fuel cost ($f_{cost}$) + Battery loss ($f_{ess}$), Lagrangian constraints & \textbf{Optimality Gap}, \textbf{Feasibility Rate}  \\
\hline

\end{tabular}
\vspace*{2mm}
\end{table*}

\cite{xu2020} compare their simulation-constraint double dueling \ac{DQN} agent with a basic double dueling \ac{DQN} agent. The simulation-constraint agent outperforms the basic agent, maintaining grid stability for longer durations per episode. They also find that agents with \ac{GNN} layers outperform those without. While results are promising, further testing on larger grids is needed to confirm the effectiveness of the approach.

The experiments in \cite{anguiano2025graph} are conducted exclusively on a very small 5-bus grid. Agents are compared only against the “Do Nothing” baseline. Results show faster convergence and higher stability when using the ElementGraph formalization with some agents achieving L2RPN scores of 100. However, the study does not include comparisons with existing GRL agents, nor does it test scalability on larger systems such as the IEEE 118 grid. Consequently, while the results demonstrate internal consistency, the practical scalability and competitiveness of the approach remain unverified, particularly given that the 5-bus can typically be solved by simple heuristics.

\cite{taha_learning_2022} trained their \ac{GNN} using features representing power lines and a reduced injection horizon for speed. To gauge topology generalization, the \ac{GNN} was tested on topologies differing by actions from the reference, showing that the RMSE increases logarithmically with action distance. Although no comparison with non-graph-based neural networks is provided, their \ac{MCTS} agent significantly reduces failure rates from 15.1\% to 1.5\%, demonstrating the effectiveness of combining \acp{GNN} for prediction and \ac{MCTS} for optimization, and paving the way for model-based \ac{RL} with \acp{GNN}.

\cite{yoon2021winning} validated their \ac{SMAAC} approach against baselines like \ac{DDQN} and \ac{SAC}, which underperformed on medium and large grids due to inefficient action exploration and potential failures. While they compared \ac{GNN}-based and non-\ac{GNN} methods, detailed validation of specific \ac{GNN} architectures was lacking. The \ac{SMAAC}/AS method, which incorporated goal topology without afterstate representation, performed poorly, highlighting the value of afterstate representation. Another baseline from the same L2RPN challenge struggled with the vast action space despite initial promise. \ac{SMAAC} significantly outperformed other methods.

In \cite{sar_multi_agent_2023}, \ac{SACD} and \ac{PPO} are evaluated in both independent and dependent multi-agent settings. Independent agents, each with its own actor, critic, and replay buffer, face coordination and stability issues. Dependent versions utilize a centralized critic for improved coordination, thereby enhancing stability and performance. \ac{SACD} performs well in a single-agent setting but is unstable in multi-agent scenarios, except for DSACD with tuned parameters. \ac{PPO} converges effectively in both single- and multi-agent settings, with faster convergence in single-agent and less sensitivity to hyperparameters in multi-agent scenarios. The \ac{GNN} used is not compared to feedforward networks or other \ac{GNN} architectures.

\cite{zhao2022graph} uses GraphSAGE networks on a modified IEEE 118-bus system, training them unsupervised and testing on various unseen grid topologies. They use 2D t-SNE to demonstrate the robust representation of the grid across different setups. Compared to a dense-based \ac{PPO} algorithm, the GraphSAGE-based method performs well even with changing grid structures, while the dense-based approach struggles to adapt effectively. The evaluation focuses on training outcomes rather than power grid performance metrics such as agent survival time.

\cite{xu_active_2022} evaluate their simulation-driven \ac{GRL} method using the L2RPN Robustness Track challenge dataset. The method, which combines decisions from sub-agents, prevents overloads more effectively than a "do-nothing" approach and achieves fast decision-making with an average time of 35 ms per step. \ac{GAT} models show better stability and economic benefits compared to \ac{GCN}, although no comparison with feedforward networks is provided. Their Long-Short-Term action deployment strategy outperforms fully reward-guided and enumeration strategies by managing overloads with fewer actions, and the action threshold of $0.98$ is validated as optimal. However, multiple action thresholds have been used across the literature, and no single value consistently dominates in terms of performance. The action threshold choice appears to be dependent on both the grid size and the utilized method \cite{van2025survey}.

The distributed GNN-based architecture proposed in \cite{fabrizio2025} is evaluated on Grid2op's IEEE 14-bus grid. Their whole system—combining GNN-based local observations, DQfD pre-training, and potential-based reward shaping—achieves over 6000 average survival steps, far exceeding the “do-nothing” baseline. Ablation studies confirm the necessity of both the GNN and imitation-learning components. Furthermore, inference time is an order of magnitude faster than the simulation-based expert used for demonstration generation, indicating the model’s potential for scalability and real-time applicability. Nevertheless, despite its conceptual scalability, the evaluation remains confined to a small 14-bus grid. This limitation is particularly relevant given that the 14-bus grid can often be solved even by a simple greedy baseline. Consequently, the claimed transferability and scalability of the approach to larger grids remain untested, and further validation on larger IEEE systems or real-world networks would be required to substantiate these claims. Moreover, the study does not report the quantitative performance of the expert system used to generate demonstrations, nor does it include comparisons with other published agents. As a result, while the proposed framework demonstrates strong internal consistency, its relative performance against state-of-the-art baselines remains unclear.

Looking at the evaluation of \cite{peter2025dualpolicy}, all simulations are conducted on a modified IEEE 14-bus grid with an opponent increasing system load by $25\%$. The study reports average survival times across $N$–$k$ contingency scenarios ($k=1\!-\!5$) and compare it to a baseline without any agent. While the dual-policy agent significantly extends survival under these conditions, no comparisons with existing RL or heuristic agents are provided, and the scalability to larger grids (e.g., IEEE 118) remains untested. However, their dual policy agents achieve comparable performance for contingency scenarios that go beyond $k=1$, making it a very promising approach. Noteworthy, however, is the limitation to just 100 simulation steps per episode, which constitutes only a fraction of the full operational horizon typically evaluated in Grid2Op. As a result, the reported survival times and robustness metrics capture only short-term resilience rather than sustained stability over realistic time spans. 

\cite{wu_constrained_2023} evaluated their \ac{GRL} approach for managing \ac{BESS} against baseline methods such as \ac{DQN} and \ac{DDPG}. They compare their spatio-temporal \ac{GCN} (Cplx-\ac{STGCN}) with feedforward, convolutional, and recurrent networks, highlighting their effectiveness. The study also tests hybrid \ac{OPF} solvers, DeepOPF, and DC3, and compares them with \ac{RL} methods like TD3, assessing metrics such as testing rewards and control over power generation and voltage magnitude. Their constrained \ac{GRL} framework outperforms traditional optimization and existing \ac{RL} techniques.

\subsection{Discussion}
\label{sec:TG_discussion}
The reviewed works demonstrate the significant potential of Graph Reinforcement Learning for transmission grid control. However, a critical analysis reveals several shared limitations and methodological gaps that must be addressed for these approaches to move from proof-of-concept to real-world applicability. These challenges can be grouped into limitations of the simulation environment, the need for standardized evaluation, and open research questions in model design.

\paragraph{The Simulation-to-Reality Gap}
Almost all approaches rely on the Grid2Op framework \cite{grid2op} for training and testing. While transmission system operators have designed it, it abstracts from real-world grid aspects. There is a need for larger grids with real injection data, realistic failure handling (e.g., N-1 security), and more accurate modeling of substation constraints and operational practices \cite{rl2grid}. Consequently, rewards, actions, states, and graph representations are limited to the functionalities and data provided. As a result, the presented approaches are, to some extent, tailored to the specific problem setup used in Grid2op. This customization means the approaches inherit the same abstractions and limitations, questioning their applicability to real-world grids. 

In particular, several \textbf{concrete pitfalls} arise from the reliance on simplified IEEE test cases and the abstractions within Grid2Op. Most studies evaluate their methods on comparatively small grids, which remain significantly below the size and complexity of real transmission networks. The commonly used double-busbar abstraction captures only part of the complexity of real-world substations. Furthermore, most benchmarks rely on synthetic injection data, as real-world measurements with realistic temporal dynamics are rarely available for research use. Another limitation is that most benchmark tasks neglect temporal N-1 security—evaluating safety only for single-step contingencies—whereas real operations require the grid to remain secure for extended horizons following any equipment failure. Grid2Op approximates N-1 assessments through its adversarial opponent mechanism, which induces random line disconnections to emulate unforeseen contingencies. However, this adversarial setup does not constitute a full N-1 evaluation, as it captures only isolated failures and lacks the temporal dimension and systematic coverage required for real-world reliability assessment. Additionally, the stochastic elements in Grid2Op—such as opponent events or random maintenance schedules can lead to non-reproducible comparisons when random seeds and horizons are not standardized. It is important to note, however, that these limitations do not only originate from the Grid2Op framework itself. Grid2Op is technically capable of handling large-scale and realistic grids through custom solvers such as LightSim2Grid or PowSyBl integrations—and even supports full N-1 security evaluations with minor configuration changes. The primary bottleneck is instead the availability of open, high-fidelity grid and injection data. Recent initiatives such as RTE’s open-source 7000 Nodes Dataset \cite{rte_os_grid}, which provides the complete French transmission network in node-breaker topology, mark a crucial step toward addressing this gap.

\paragraph{The Need for Standardized Evaluation}
To mitigate these limitations, \textbf{standardized evaluation protocols} are required. Future studies should use consistent evaluation horizons and report stochastic seeds. Moreover, benchmarks should specify the level of stochasticity, contingency modeling, and grid size to ensure comparability across publications. Some studies provide in-depth evaluations of \ac{RL} algorithms and architectures and propose averaging results over multiple random seeds to increase reliability \cite{binbinchen,hassouna2025_soft,lehna2024fault}. Based on their evaluation setups, using 20 to 30 seeds appears to establish a reliable basis for comparison.

While evaluation on real grid injection data is the ultimate goal, utilizing widely available synthetic data remains a viable option for benchmarking methods. Such evaluation environments should be of sufficient size; we follow the example set by \cite{lehna2024hugo} and recommend a year's worth of data (i.e., 52 weeks/chronics).
Regarding the evaluation, the corresponding metrics should align with the objectives that real-world grid operators use in their decision-making. These objectives are \cite{gridoptions}: 
\begin{itemize}
    \item Minimizing the N-1 load flow
    \item Minimize the number of switching actions
    \item Minimize the number of open busbar couplers
    \item Minimize topology distance, i.e., minimize the amount of switching when stepping from one topology to another
\end{itemize}

\paragraph{Need for Consolidation}
Beyond evaluation, the primary methodological gap appears to be a need for consolidation. The research landscape is characterized by numerous, isolated innovations. For instance, sophisticated graph representations, such as heterogeneous graphs \cite{dejong2025}, have been proposed but not yet combined with advanced algorithmic structures, like afterstate representations \cite{yoon2021winning}. Likewise, specialized \ac{GNN} architectures (such as \cite{wu_constrained_2023}) and planning techniques (such as \ac{MCTS} \cite{taha_learning_2022}) have been developed in parallel but have not yet been integrated into unified frameworks. A significant opportunity exists in synthesizing these innovations—for example, combining a heterogeneous \ac{GNN} with a hierarchical agent and an afterstate representation.

In addition to this need for consolidation, several other promising research avenues remain largely underexplored. These include the deep, systematic design of graph representations, the development of specialized \ac{GNN} architectures, and the crucial work of mapping grid problems to the most appropriate \ac{RL} algorithms—particularly hybrid approaches combining imitation learning with \ac{GRL}.

\paragraph{Open Frontiers in Graph Neural Network design}
Most studies use similar state features but differ in the information included, demonstrating the flexibility of \ac{GRL}. However, despite the clear performance advantage of using graph representations, the studies lack a deep, systematic analysis of how grids are optimally represented. Recent work in imitation learning, for example, has demonstrated that heterogeneous graph models (which explicitly model different connection types, such as "same-busbar" vs. "cross-busbar"), can overcome information asymmetry and improve generalization \cite{dejong2025}.  More sophisticated graph modeling, such as hyper-heterogeneous multi-graphs \cite{Donon2024}, has not yet been incorporated into \ac{GRL} approaches. 

\acp{GNN} are pivotal for extracting features from power grids, enhancing convergence and generalizability across different configurations.  While \acp{GAT}, GraphSAGE, and transformers are used, their evaluation against alternatives is often shallow. Furthermore, most applied architectures are generic; specialized architectures, such as the Cplx-\ac{STGCN} \cite{wu_constrained_2023}, are the exception. There is a clear opportunity to design \ac{GNN} architectures specifically adapted to power grid physics and topologies, moving from graph-level predictions of single actions toward richer, node-level predictions of the entire grid state \cite{dejong2025}. The novel and power grid-specific architecture described in Sec.~\ref{gnn_architectures} would be a promising pathway.

\paragraph{Mapping Problems to RL Algorithms}

The diversity in \ac{RL} methods highlights a critical gap: the lack of a clear mapping between specific grid problems and the most suitable algorithmic family. This review suggests the following trade-offs.

\textbf{Topological Control} This problem is defined by its vast, discrete, and combinatorial action space. Standard model-free methods, and in particular on-policy RL like \ac{PPO} are prone to face a severe failures when directly applied, suffering from poor sample efficiency as they struggle to explore the long-term consequences of complex switching sequences.

\textbf{Addressing Topological Complexity} The "afterstate" methods \cite{yoon2021winning}, hierarchical/multi-agent decompositions \cite{sar_multi_agent_2023, fabrizio2025}, and action masking \cite{anguiano2025graph} are all direct responses to this challenge, designed to prune the action space. MCTS-assisted selection \cite{taha_learning_2022, xu_active_2022} is another solution, trading a degree of sample efficiency (e.g., requiring a simulator in the loop) for more robust, safety-verified planning, though this can be computationally expensive at inference. However, model-based techniques, such as those in \cite{enlite}, are rarely used, and no full model-based \ac{GRL} agent has yet been developed. Moreover, advanced model-free methods such as BBF \cite{bbf} that could improve sample efficiency have not been applied.

A key underexplored strategy is the combination of imitation learning (IL) with \ac{GRL}. IL is a proven method to accelerate training and improve sample efficiency by overcoming the "cold start" problem \cite{binbinchen, dejong2025, curriculumagent}. The success of \ac{GNN}-based IL methods using rich supervision signals, such as soft labels \cite{hassouna2025_soft}, suggests that a hybrid IL-\ac{GRL} agent could offer a powerful balance of expert-guided safety and \ac{RL}-driven optimization.

\textbf{Redispatch and Storage Control} In contrast, problems like generator redispatch or battery storage \cite{zhao2022graph, wu_constrained_2023}, which often involve continuous or smaller discrete action spaces, are more amenable to constrained \ac{RL} or standard model-free algorithms. Furthermore, integrating topological actions with generator redispatch is crucial for a more flexible and cost-effective control strategy; agents can leverage near-zero-cost substation reconfigurations to alleviate the majority of congestion, resorting to expensive generation adjustments only for remaining overloads.

To sum up, while the proposed approaches demonstrate significant potential, they remain largely at the proof-of-concept stage and are not yet scalable to real-world operations. Future deployment hinges not just on performance, but on ensuring trustworthiness and transparency \cite{mussi2025}. This necessitates a paradigm shift toward human-AI decision support frameworks rather than full automation \cite{leyliabadi2025}. Nevertheless, these works successfully pave the technical foundation for next-generation grid operation.

\section{Graph Reinforcement Learning for Distribution Grids}
\label{chap:DG}

Power generation has increasingly shifted from the transmission system to the distribution side \cite{beinert_power_2023} due to the rise of distributed renewable energy sources such as photovoltaics. This shift causes voltage fluctuations \cite{srivastava_voltage_2023} that can threaten grid stability, as system voltages must remain within operational limits. Voltage control addresses these issues by flattening voltage profiles and reducing network losses using devices such as voltage regulators, switchable capacitors, and controllable batteries \cite{fan2022powergym}, as well as topology control \cite{xu_online_2022}. \ac{RL} is particularly promising for handling multiple objectives in voltage control optimization problems. While \ac{DRL} has shown promise in this area \cite{duan2020_rl_survey}, the combination with \acp{GNN} is still emerging. We note that \ac{DRL} methods serve as a proof of concept and are still far from practical deployment.

Classical methods for distribution grid tasks, like OPF-based optimization, mixed-integer formulations for discrete switching, and rule- or heuristic-based control, remain the standard toolkit in practice. However, they face critical limitations in modern distribution networks, which are increasingly dominated by distributed energy resources \cite{yang2023opfreview}. Similarly to transmission grid operation, accurate AC behavior combined with discrete actuators (tap steps, capacitor switching, feeder reconfiguration, shedding) leads to nonconvex mixed-integer programs that are difficult to solve reliably within real-time deadlines \cite{heid2025}. Linearizations and convex relaxations lose exactness under high R/X ratios, tight voltage bounds, or temporary meshing conditions common in distribution networks \cite{pham2024,liao2019}. Combinatorial action spaces in reconfiguration and restoration scale poorly, which often necessitates heuristics that trade optimality for robustness \cite{lotfi2024}. Many formulations also assume full observability and stable parameters, making them sensitive to missing or noisy measurements, topology uncertainty, and forecast errors \cite{liao2019}.

In contrast, GRL learns state-to-action policies over the grid graph and can naturally handle the given challenges, which is explored in the remainder of this chapter. An overview of the logical flow in \ac{GRL} for grid control can be found in Fig.~\ref{fig:grloverview}. 

\begin{table*}[!htbp]
\centering
\renewcommand*{\arraystretch}{1.4}
\caption{\textbf{Overview of \ac{GRL} approaches for distribution grids under \textit{operational control}}. The \textit{action} column lists devices or variables modified by predicted actions (\textit{q} for reactive and \textit{p} for active power, ESS for energy storage systems, SVC for static var compressors). The column \textit{Grid size} lists the number of grid buses, and  \textit{Focus/Unique Feature} highlights key aspects or major differences to other approaches.}
\label{table:distribution_grid_operation}
\begin{tabular}{|M{0.5cm}||P{1.9cm}|P{2.0cm}|P{2.5cm}|P{2.0cm}|P{1.2cm}|P{3.1cm}|}
\hline
& {Approach}           & {RL }     & {Action}          & {GNN}         & {Grid Size}           & {Focus/ Unique Feature} \\ 
\hline\hline

\multirow{10}{*}[-4.5cm]{\rotatebox[origin=c]{90}{Operational Control}} 
& Yan et al. 2023 \cite{yan_multi_2023} & MAAC & q (PV inverters) & Spectral GCN & 141 & Zoned grids, primal–dual approach \\ \cline{2-7}
& Mu et al. 2023\cite{mu2023graph} & MAAC & q (PV inverters) & Spectral GCN & 141, 39 & Zoned grids \\ \cline{2-7}
& Hu et al. 2024 \cite{hu2023multiagent} & MASAC & q (PV, SVC)& Hierarchical spatio-temporal GAT & 33, 123 & Fully decentralized training (DTDE), partial observability, robustness to communication failure \\ \cline{2-7}
& Wang et al. 2023 \cite{wang2023graph} & Multi-agent PPO & ESS, p (generators) & Spatial GCN & 69, 123 & Real-time operation, multiple microgrids \\ \cline{2-7}
& Lee et al. 2022 \cite{lee2022graph} & PPO & ESS, voltage regulators, capacitors & GAT & 13, 34, 123, 8500 & Graph augmentation, local readout \\ \cline{2-7}
& Wu et al. 2023 \cite{wu2023two} & AC & q (PV inverters) & Custom GSO (spectral) & 33, 25 & Two-stage hybrid (optimizer + RL), grid-specific filter \\ \cline{2-7}
& Wu et al. 2022 \cite{wu_reinforcement_2022} & PPO & q (PV inverters) & Custom GSO (spectral) & 33, 119 & Grid-specific filter in GNN \\ \cline{2-7}
& Cao et al. 2023 \cite{cao_physics-informed_2023} & AC & q (PV inverters), ESS, var compressors & GAT & 33 & Surrogate GNN (grid embedding + reward) \\ \cline{2-7}
& Li et al. 2023 \cite{li2023deep} & PPO & p (generators) & Spatio-temporal GAT & 33, 69, 118 & Consider temporal information \\ \cline{2-7}
& Xing et al. 2023 \cite{xing_real-time_2023} & DDPG & p (generators, PV, flexible loads), q (SVC) & GAT & 33, 119 & Computational efficiency, multiple objectives \\ \cline{2-7}
& Xu et al. 2022 \cite{xu_online_2022} & DQN & Topo & Spectral GCN & 33, 69, 118 & Action space reduction via GNN + branch exchange \\ 
\hline
\end{tabular}
\end{table*}

\begin{table*}[!htbp]
\centering
\renewcommand*{\arraystretch}{1.4}
\caption{\textbf{Overview of \ac{GRL} approaches for distribution grids under \textit{emergency mode}}. The \textit{action} column lists the actions available to the agent. The column \textit{Grid size} lists the number of grid buses and  \textit{Focus/Unique Feature} highlights key aspects or major differences to other approaches.}
\label{table:distribution_grid_emergency}
\begin{tabular}{|M{0.5cm}||P{1.9cm}|P{2.0cm}|P{2.5cm}|P{2.0cm}|P{1.2cm}|P{3.1cm}|}
\hline
& {Approach}           & {RL }     & {Action}          & {GNN}         & {Grid Size}           & {Focus/ Unique Feature} \\ 
\hline\hline

\multirow{4}{*}[-1cm]{\rotatebox[origin=c]{90}{Emergency Mode}} 
& Hossain et al. 2021 \cite{hossain_graph_2021} & Double DQN & Binary load shedding & Spatial GCN & 39 & Consider temporal information \\ \cline{2-7}
\cline{2-7}
& Pei et al. 2023 \cite{pei2023emergency} & Double-Dueling DQN & Two-level load shedding & GraphSAGE & 39, 300 & Adaptability to unseen topologies \\ \cline{2-7}
& Zhao et al. 2022 \cite{zhao_learning_2022} & Q-Learning & Reconnection of grid components & GCN & 123, 8500 & System restoration \\ \cline{2-7}
& Jacob et al. 2024 \cite{jacob2024real} & PPO & load shedding, line switching  & Graph
capsule
network & 13, 34, 123 & achieves a close to optimal compliance with constraints  \\ 
\hline
\end{tabular}
\end{table*}

\subsection{Voltage Control}
Voltage control is the key task in the distribution grid, managing reactive power set points to maintain grid stability. While active power is the power that runs devices, reactive power is required to provide the voltage levels that enable the delivery of real power. The cost function in these tasks typically includes system-wide indicators such as power losses and congestion \cite{srivastava_voltage_2023}. It is essential that the voltages remain within the prescribed limits, as any violation would have detrimental effects on the system.

The AC power flow in a grid is modeled by highly non-linear equations, making the optimization problem non-convex. To simplify, it is often linearized using methods like DC-OPF formulations \cite{DCOPF}. For exact models, numerical methods such as Newton-Raphson or Gauss-Seidel are used. Heuristics, such as particle swarm optimization, address the non-convexity of the problem formulation \cite{srivastava_voltage_2023}. Deep learning approaches offer an effective alternative since they are suited for non-linear problems and overcome the limitations of traditional methods in dealing with the complexity and dynamics of smart grids.

In this work, we distinguish between two cases of voltage control: operation control and emergency mode. In the case of operation control, i.e. a stable grid state, voltage control is typically addressed through reactive power control. Emergency situations require more drastic measures such as load shedding, i.e., the cutting of loads to prevent the grid from blackout. Tab.\ref{table:distribution_grid_operation} and ~\ref{table:distribution_grid_emergency} list the \ac{RL} method, action type, GNN architecture, grid size, and overall focus of the analyzed approaches.

\subsubsection{Operational Control.}
A commonly used approach to managing bus voltages is through the control of the reactive power which enables the generation of electromagnetic fields without delivering usable power to consumers. Modern inverter-based generation and digitized grids can maintain voltage within the desired range through reactive power control, aiming to minimize network loss, mitigate voltage oscillations, and reduce operational costs.\\

\paragraph{\normalsize{Reinforcement Learning Framework}}

\paragraph{Rewards.}
Grid stability relies heavily on maintaining voltages within defined limits, so all approaches penalize voltage deviations from a reference value. Many methods combine penalties for voltage deviation with other terms such as power loss \cite{yan_multi_2023, mu2023graph}, equipment wear \cite{lee2022graph}, and voltage barrier functions to constrain voltage ranges \cite{mu2023graph}. Table \ref{tab:distribution_metrics} provides a comprehensive summary of these reward formulations. Terms for PV curtailment, voltage oscillation \cite{wu_reinforcement_2022, li2023deep}, renewable integration, and generation costs \cite{li2023deep} are also considered. \cite{cao_physics-informed_2023} and \cite{xu_online_2022} also base their reward on voltage deviation. But in addition, \cite{cao_physics-informed_2023} introduce a surrogate model that estimates voltage and power loss based on the state and action and \cite{xu_online_2022} penalize actions on already disconnected lines and define rewards based on the tree metric basic cut set to avoid loops or branch disconnections. In general, balancing multiple reward components is critical, as optimizing one can negatively impact others.

\paragraph{Actions.}
Voltage regulation mostly involves adjusting actuator set points. Approaches like \cite{yan_multi_2023, mu2023graph,wu2023two, wu_reinforcement_2022, cao_physics-informed_2023, xing_real-time_2023} adjust reactive power outputs of PV inverters. while others also control the active power of \ac{ESS} \cite{lee2022graph, wang2023graph, cao_physics-informed_2023,xing_real-time_2023}, flexible loads \cite{xing_real-time_2023} and static var compensation \cite{xing_real-time_2023, cao_physics-informed_2023}. Adjusting the active power of generators of renewables (\cite{wang2023graph,li2023deep}) or diesel generators (\cite{wang2023graph}) alongside reactive power is another option. In contrast, \cite{xu_online_2022} focuses on modifying grid topology by disconnecting and reconnecting lines.

\paragraph{States.}
The presented \ac{GRL} methods predict actions based on states such as voltage measurements, load demand, and power generation. The status of actuators, e.g., \ac{ESS}, tap changers \cite{xu_online_2022,lee2022graph, wu2023two}, and electricity grid prices \cite{wang2023graph} are also considered. States are typically embedded using \acp{GNN} to encode the distribution grid's features and topology. These encoded representations are then used by the \ac{RL} algorithm for decision-making.

\paragraph{Reinforcement Learning Algorithms}\
Common \ac{RL} algorithms in these studies include Actor-Critic, \ac{PPO}, and Q-Learning, each tailored to specific setups and objectives. Approaches like \cite{lee2022graph, wu_reinforcement_2022, xing_real-time_2023, li2023deep} use a \ac{GCN} for grid embedding and train policies with \ac{DDPG} or \ac{PPO}. Multi-agent actor-critic setups \cite{yan_multi_2023, mu2023graph} with one agent per zone manage zoned networks with centralized training including global observations and decentralized evaluation, i.e., based only on local observations. \cite{yan_multi_2023} integrate the \ac{GNN} into the actor networks, while \cite{mu2023graph} use the \ac{GCN} only in the critic to model agent interactions. To ensure that the physical equations of the physical system are satisfied, a primal-dual method similar to that of \cite{wu_constrained_2023}, as described in Sec.~\ref{subsec:TG_approaches}, is used.

A significant challenge in these multi-agent setups is the reliance on global information for centralized training. Addressing this, \cite{hu2023multiagent} propose a fully Decentralized Training and Decentralized Execution (DTDE) framework. Their algorithm, MASAC-HGRN, is based on Multi-Agent \ac{SAC} and uses a stochastic policy with maximum entropy to improve exploration. Unlike centralized training methods, each agent in this paradigm learns using only its local observations and information communicated from its immediate neighbors, making it more robust to the partial observability inherent in real-world systems.

Similarly, \cite{wang2023graph} use multi-agent \ac{PPO} with a \ac{GNN} in both actor and critic for microgrid management, where each microgrid is controlled by an agent that manages its power schedule.

\cite{wu2023two} propose a very different two-stage approach: day-ahead optimization of storage systems, tap changers and capacitor banks settings using Mixed Integer Second Order Cone Programming, followed by actor-critic learning for voltage regulation in a continuous action space with \ac{GCN}-based grid embeddings. Similarly, \cite{cao_physics-informed_2023} use a \ac{GCN} to embed the grid, but they apply a subsequent fully connected deep autoencoder for feature dimension reduction in an actor-critic framework. Only \cite{xu_online_2022} controls voltages by adjusting the grid topology, addressing the NP-hard complexity with a method to reduce the action space. First, they use deep Q-learning to predict a line to be disconnected. Second, they apply a branch exchange mechanism that considers the radial constraints (i.e., maintaining a tree structure without loops) when selecting a line to be reconnected. 

\paragraph{\normalsize{Graph Embeddings}}
The \ac{GNN} architectures extract features from bus-centric graphs as described in Sec.~\ref{gnn_architectures} where nodes represent buses including properties such as injections, voltage, active and reactive power and the status of controllable actuators). However, the \acp{GNN} vary across the presented approaches. Multiple approaches employ the popular \ac{GAT} (cf. Eq.~\ref{equ:att_conv}) or some variant of it and train it using an \ac{RL} algorithm. \cite{lee2022graph, li2023deep, xing_real-time_2023}. By calculating attention coefficients for each edge (as shown in Eq. \ref{equ:att_weights}), these models can prioritize critical neighbors—such as a specific bus experiencing a voltage violation—while reducing the influence of less relevant distant nodes.

Some studies \cite{wang2023graph,lee2022graph} extend their graph before applying a \ac{GNN}. \cite{lee2022graph} introduce additional edges to allow faster information propagation through the graph. For decentralized microgrid voltage control, \cite{wang2023graph} construct a graph comprising only critical buses linked to generators, microgrids, or feeder endpoints that are connected via edges weighted by electrical distance.

Several approaches apply spectral graph convolutions (cf. Eq.~\ref{equ:spec_approx}) for feature extraction \cite{yan_multi_2023,xu_online_2022, mu2023graph, wu_reinforcement_2022}. \cite{xu_online_2022} learn to predict the best line to be disconnected using three layers of graph convolutions, each followed by an MLP. The output is an edge embedding generated by aggregating the embeddings of the incident node. \cite{yan_multi_2023} employed a spectral \ac{GCN} as actor-networks of each control agent. It acts like a low pass filter that suppresses noise in the input data and  fills in missing values by aggregating the features of neighbouring nodes. As defined in Sec. \ref{subsubsec:messagepassing} (Eq. \ref{equ:spec_approx}), spectral convolutions operate by filtering graph signals in the frequency domain defined by the graph Laplacian. This property allows the \ac{GCN} to efficiently attenuate high-frequency noise variations common in sensor data while preserving global structural signals. In contrast, \cite{mu2023graph} use the \ac{GNN} only in the critic. It receives the action predicted by the actor, along with the observations, and combines them with the grid information from neighbouring agents, i.e. neighbour grid zones.

To facilitate its decentralized, multi-agent framework, \cite{hu2023multiagent} design a novel architecture, the Hierarchical Graph Recurrent Network (HGRN), to explicitly handle partial observability and agent heterogeneity. In their model, each agent (representing a grid region) is a node in a meta-graph. The HGRN architecture, which is used in both the actor and critic, consists of three stages: (1) an encoder maps heterogeneous local observations to a common embedding space; (2) a Hierarchical \ac{GAT} aggregates information from neighboring agents, allowing for communication; and (3) the resulting embedding is processed by a Gated Recurrent Unit, which maintains a memory to compensate for the partial observability of the environment.

\cite{wu_reinforcement_2022} present a spatial temporal \ac{GNN} including a graph shift operator, a spectral filter based on the relationship between voltage angle and magnitude of AC power flow equations similar to the architecture presented \cite{wu_constrained_2023} (Sec.~\ref{chap:TG}). Furthermore, the approach incorporates temporal information by aggregating the node embeddings of the last 10 time steps. \cite{wu2023two} also introduce a graph shift operator but it is based on the grid topology and the correlation coefficient matrix obtained from the PV and historical load data. Both \cite{wu2023two} and \cite{wu_reinforcement_2022} use one convolutional layer and an MLP-based readout network comprising three layers.
Another  spatio-temporal approach is presented by \cite{li2023deep}. Their attention mechanism learns the temporal dependency of a node's embedding at different time steps. The convolutions are applied sequentially to combine spatial and temporal information. To account for different periodic patterns, a fully connected layer combines three temporal resolutions (32 hours, 16 days, 4 weeks) for the final prediction. The architecture is applied in both the actor and critic networks.
In terms of training, \cite{cao_physics-informed_2023} present an alternative to directly learning the \ac{GNN} weights using an \ac{RL} algorithm. They first train a surrogate \ac{GCN} in a supervised manner on historical power flow data to predict node voltages like the approaches mentioned in Sec.~\ref{gnn_architectures}. Then, the weights are copied to the representation networks of the actor-critic networks to perform feature extraction from the distribution network.
\begin{figure*}[htbp]
	\centering
	\includegraphics[trim=0 0 2cm 0, clip, width=\textwidth]{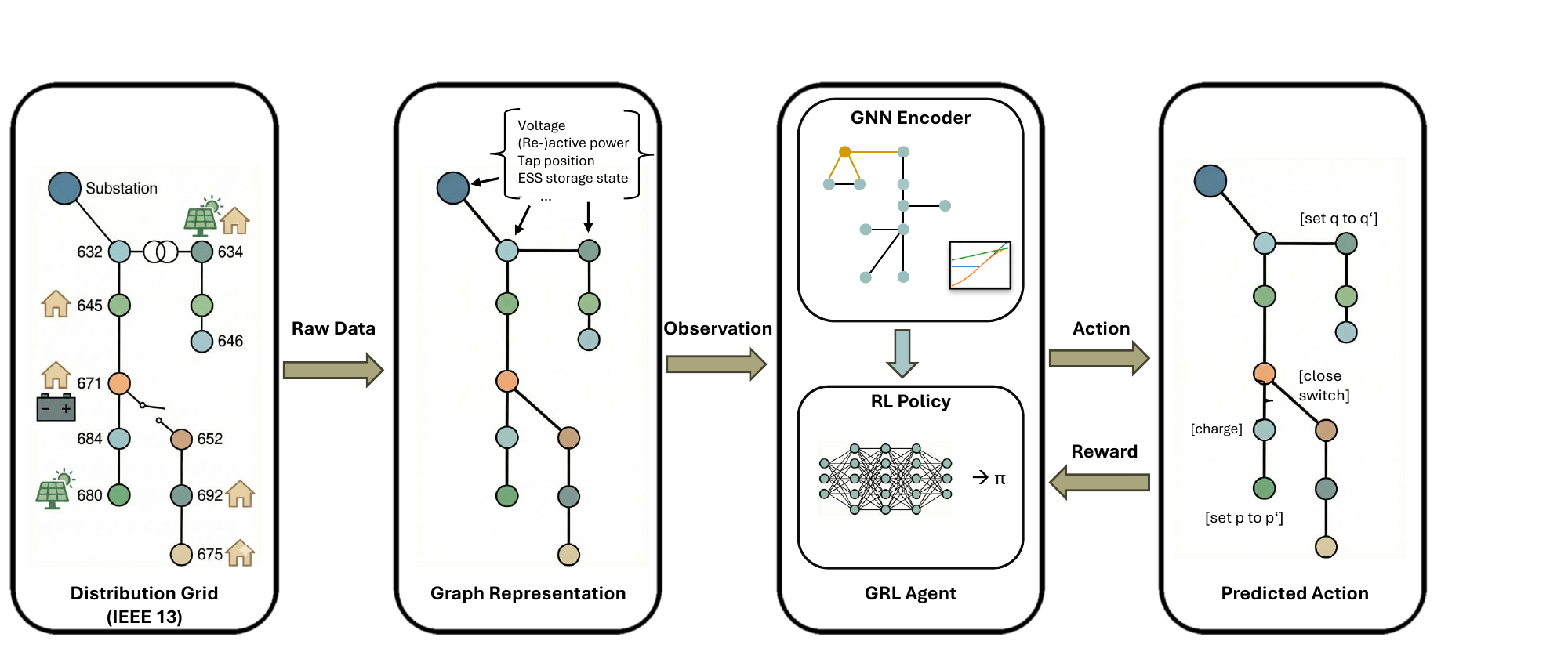}
	\caption{\textbf{Overview of the GRL approach for Distribution Grid Control.} The environment represents the distribution grid, comprising various components such as distributed energy resources and energy storage systems connected via power lines, transformers, and switches. The left side shows a schematic of the IEEE 13-node test grid, which serves as the experimental use case in some of the presented approaches. This physical system is modeled as a graph, where nodes are characterized by specific features. This graph structure is passed to the GRL agent, which utilizes an \ac{GNN} encoder to generate a comprehensive graph observation. An RL policy (typically Actor-Critic) then processes these embeddings to predict specific control actions, such as reactive power adjustments or tap position changes. These actions are executed within the grid environment, and a reward signal—calculated based on defined grid metrics, such as minimizing voltage deviation or power loss—is returned to the agent to update its weights.}
 \label{fig:dg_overview}
\end{figure*}
\paragraph{\normalsize{Experiments and Evaluation}}
Most approaches evaluate their performance on IEEE grids ranging in size from 5 to 300 buses (see \ref{table:distribution_grid_operation}). The data for the injections can be randomly sampled or correspond to historical time series of power generation, mostly from photovoltaic. The authors evaluate the presented approaches using metrics such as voltage deviation, network energy loss, or voltage violation rates. Comparisons typically include traditional optimization methods, heuristics, and other deep \ac{RL} approaches as benchmarks. Table \ref{tab:distribution_metrics} details the specific reward components used to guide the agents and the key performance metrics (e.g., robustness to noise, inference speed, violation rates) used to validate efficacy against baselines.

\begin{table*}[!htb]
\centering
\small
\caption{Overview of Reward Variables and Experimental Metrics (Distribution Grid)}
\label{tab:distribution_metrics}
\renewcommand{\arraystretch}{1.2}
\begin{tabular}{|M{3.5cm}||P{5.5cm}|P{5.5cm}|}

\hline
\textbf{Reference}
& \textbf{Reward Variables (Optimization Objective)}
& \textbf{Experimental Performance Metrics}
\\
\hline
\hline
\multicolumn{3}{|c|}{\textbf{Operational Control} (\textit{Focus: Voltage stability, loss minimization, and economic dispatch})} \\
\hline
Yan et al. 2023 \cite{yan_multi_2023} & Voltage deviation penalty, Active power loss, Lagrangian safety constraints & Energy loss reduction, Voltage violation rate \\
\hline
Mu et al. 2023 \cite{mu2023graph} & Voltage deviation, Active power loss, Barrier function for limits & Robustness to \textbf{topology changes} (N-1 contingencies) \\
\hline
Hu et al. 2024 \cite{hu2023multiagent} & Power loss, Voltage deviation penalty & Performance maintenance under \textbf{communication failure} \\
\hline
Wang et al. 2023 \cite{wang2023graph} & Microgrid power schedule costs, Curtailment penalty & Scalability to larger grids, Near-optimal performance \\
\hline
Lee et al. 2022 \cite{lee2022graph} & Voltage deviation, Switching cost (Equipment wear) & Robustness to \textbf{noisy/missing sensor data} \\
\hline
Wu et al. 2023 \cite{wu2023two} & Voltage deviation, Active power loss, Smoothness penalty & Voltage profile improvement, Loss reduction vs. DDPG \\
\hline
Wu et al. 2022 \cite{wu_reinforcement_2022} & Voltage oscillation penalty (Frequency stability) & Mitigation rate of compromised inverters (Cyber-attack) \\
\hline
Cao et al. 2023 \cite{cao_physics-informed_2023} & Voltage deviation, Surrogate model estimation error & Voltage deviation reduction, Training acceleration \\
\hline
Li et al. 2023 \cite{li2023deep} & Voltage oscillation, Generation costs, Renewable usage & Inference speed, Reward convergence \\
\hline
Xing et al. 2023 \cite{xing_real-time_2023} & Multi-objective: Voltage, Loss, and Comfort (Flexible load) & Computational efficiency, Pareto optimality \\
\hline
Xu et al. 2022 \cite{xu_online_2022} & Voltage deviation, Tree metric (penalize loops/islands) & Inference speed vs Heuristics, Optimality gap \\
\hline
\hline

\multicolumn{3}{|c|}{\textbf{Emergency Mode} (\textit{Focus: Load shedding and system restoration})} \\
\hline
Hossain et al. 2021 \cite{hossain_graph_2021} & Binary Voltage Recovery (+10/-10), Stability penalty & \textbf{Convergence rate} (Episodes to stability), Recovery time \\
\hline
Pei et al. 2023 \cite{pei2023emergency} & Minimize total load shedding magnitude ($P_{shed}$) & Adaptability to \textbf{unseen fault locations} \\
\hline
Zhao et al. 2022 \cite{zhao_learning_2022} & Restoration steps (minimize), Reconnection success & Inference speed vs CPLEX \\
\hline
Jacob et al. 2024 \cite{jacob2024real} & Maximize power supply, Penalize non-convergence & \textbf{Constraint compliance rate}, Optimality gap \\
\hline

\end{tabular}
\vspace*{2mm}
\end{table*}
The advantage of \acp{GNN} over dense-based \ac{RL} agents becomes evident in several studies. \cite{lee2022graph} tested their \ac{GNN}-based \ac{PPO} on PowerGym grids ranging from 13 to 8500 nodes, showing better performance and robustness, especially with noisy and missing data. In addition, the paper finds that voltage regulators affect the grid globally, while batteries and capacitors have local effects. To address this, the authors add edges between nodes with voltage regulators and use a local readout function for the controllable nodes, improving the robustness and performance of the \ac{GNN}-based \ac{PPO} approach. Similarly, \cite{wang2023graph} compared their graph \ac{PPO} with a dense \ac{PPO} method on IEEE grids from 33 to 123 buses, demonstrating near-optimal performance and better scalability than other multi-agent \ac{RL} methods as well as optimisation-based approaches.

Similar grid sizes are addressed by \cite{li2023deep}, whose \ac{GRL} approach significantly outperforms optimization-based benchmarks with faster inference, higher rewards, lower voltage fluctuations, and greater renewable energy accommodation. Their spatio-temporal attention exhibits a faster convergence with the attention masks, indicating strong connections to high-power buses.

\cite{yan_multi_2023} found that their primal-dual \ac{GRL} model minimized energy losses and voltage deviations and outperformed single-agent and multi-agent \ac{DDPG} methods, especially with noisy data. Similarly, \cite{mu2023graph} reported robustness to line and bus deletions as well as stable voltages and fewer violations on 33-bus and 141-bus grids when compared to optimization methods and multi-agent \ac{DDPG}. This is the same for \cite{wu2023two}, whose approach improved voltage profiles and reduced power losses on IEEE 33-node and 25-node systems compared to dense-based and \ac{CNN}-based \ac{DDPG} and conventional optimization.

\cite{wu_reinforcement_2022} mitigated oscillations from cyber-attacks in 33 and 119-node systems, showing effective mitigation even with 50\% of inverters compromised. They did not benchmark against other methods, hence, these results are difficult to interpret. Particularly, the utilization of \acp{GNN} remains to be validated. Similarly, \cite{xu_online_2022} demonstrated that their method was faster than the heuristics and close to optimal. They confirmed the benefits of \acp{GCN}, branch exchange, and action separation in an ablation study.

\cite{hu2023multiagent} validate their MASAC-HGRN algorithm on the IEEE 33-bus and 123-bus systems, comparing it against a model-based SOCP benchmark as well as centralized (DQN) and centralized-training (MADDPG, MAAC) \ac{RL} baselines. The decentralized (DTDE) approach outperformed all other \ac{RL} methods in minimizing both power loss and voltage deviation. While all \ac{RL} methods were orders of magnitude faster at inference than the SOCP solver, the key finding was in robustness. When communication links to agents were severed, the centralized DQN failed, and the CTDE-based MAAC degraded significantly. In contrast, the proposed DTDE approach showed the least performance drop, highlighting the resilience of a decentralized framework that does not depend on a central controller.

Lastly, \cite{cao_physics-informed_2023} evaluated their physics-informed \ac{GAT}-\ac{SAC} on IEEE 33- and 119-node systems. Their approach outperforms other methods several control methods, including \ac{SAC} variants and \ac{GCN}-\ac{SAC} in reducing voltage deviations and maintaining safe voltage levels, especially under noisy conditions. Ablation studies emphasized the importance of the \ac{GAT}-based network and the added robustness from the deep autoencoder. Tests on the IEEE 119-node system confirmed the method's scalability and effectiveness in larger networks.

\subsubsection{Distribution Grid Control in Emergency Mode}
Another way to control grid voltages is load shedding, which involves deliberately disconnecting certain loads. This drastic measure is usually a last resort to avoid total blackouts. Conversely, in the event of a full or partial blackout occurs, system restoration is required to restart the grid. This involves coordinated steps to reconnect loads, restore generation capacity, and ensure the integrity of distribution networks. 

\paragraph{\normalsize{Load Shedding}}
The outlined approaches \cite{hossain_graph_2021,pei2023emergency,jacob2024real} present \ac{GRL} agents for load shedding decision-making problems.
\paragraph{\normalsize{Reinforcement Learning Framework}}
\paragraph{States and Actions}
The problem is framed as an \ac{MDP} based on grid state observations such as power demand, generation, voltage measurements, and topology. \cite{pei2023emergency} also use historical node voltages. \cite{hossain_graph_2021} and \cite{jacob2024real} model decisions as binary (shed or not), while \cite{pei2023emergency} consider shedding 5\% or 10\% of the load. All approaches consider only heavy load nodes which significantly reduces the action space. \cite{jacob2024real} additionally include line switching.
\paragraph{Reward}
All approaches reward stable voltage levels and prevention of system collapse (cf. Table \ref{tab:distribution_metrics}). For example, \cite{hossain_graph_2021} give a large negative reward if the voltage has not returned to nominal within a specified time, and a positive reward if the voltages stay within predefined levels. The reward either minimizes load shedding (\cite{hossain_graph_2021, pei2023emergency}) or maximizes power supply (\cite{jacob2024real}) and penalizes actions that violate system constraints. \cite{jacob2024real} introduce a high penalization when the load flow does not converge for the predicted grid configuration to avoid invalid grid states.
\paragraph{Reinforcement Learning Algorithm}
\cite{hossain_graph_2021} and \cite{pei2023emergency} employ a \ac{DDQN} with an $\epsilon$-greedy strategy and experience replay with the latter also integrating a dueling architecture. In contrast, \cite{jacob2024real} adopt \ac{PPO} using a hybrid policy network combining a fully connected network and a \ac{GCN}. 
\paragraph{\normalsize Graph Embedding}
All studies model the grid as a graph with nodes representing buses, including substations, loads, and generators. Node features include grid state data such as voltage measurements and loads. The edges represent power lines and transformers. All approaches use a \ac{GCN} for feature extraction but with different architectures. \cite{hossain_graph_2021} use a simple \ac{GCN} with three layers followed by a narrow fully connected layer processing the flattened embeddings. \cite{jacob2024real} employ a graph capsule network (cf. Sec. \ref{chap:basics}) to embed entire graphs.
To reduce information loss in the convolution, higher-order statistics for each feature within the local neighbourhood of a node are stored along with compressed node labels, similar to Weisfeiler-Lehman colourings \cite{verma2018graph}. Unlike standard message passing that aggregates features into a single scalar or vector (Eq. \ref{equ:msgpassing}), the capsule function described in Sec. \ref{subsubsec:messagepassing} maps node features to vector representations that explicitly preserve feature statistics and orientation information, ensuring better reconstruction of local subgraphs. In the end, global grid information is integrated through a feedforward network. \cite{pei2023emergency} apply GraphSAGE, which samples and aggregates the local neighborhood of nodes.
\paragraph{\normalsize Experiments and Evaluation}
\cite{hossain_graph_2021} and \cite{pei2023emergency} use the IEEE 39-bus system, while \cite{jacob2024real} test on modified IEEE 13-bus and 34-bus systems. Each method trains on different topological configurations with random fault locations. All three approaches require less load shedding than using a fully connected network and outperform it in terms of convergence and average reward, especially on unseen topologies. It should be noted that none of the approaches is evaluated on completely new topologies as fault locations are randomly inserted in a known set of nodes. \cite{hossain_graph_2021} test on 32 topologies in the IEEE 39-bus system, while \cite{pei2023emergency} also apply their \ac{GNN} method to a 300-bus system, effectively handling larger action spaces. \cite{jacob2024real} and \cite{pei2023emergency} also outperform traditional optimization techniques in terms of speed and near-optimal performance. \cite{pei2023emergency} show that GraphSAGE is more adaptable and efficient than classical \ac{GCN}, but a comparison to a graph capsule network has not been made. The other two approaches do not compare to other \ac{GNN} architectures.

\paragraph{\normalsize{System Restoration}}
If earlier mitigation actions fail and a blackout occurs, rapid system restoration is crucial to reconnect loads and restart the grid promptly. \cite{zhao_learning_2022} employ a multi-agent approach where a Q-network guides actions by using an encoder that takes observations such as generator capacities, and load conditions as input. A \ac{GCN} extracts features from neighboring agents and passes the embeddings to the Q-networks. This method outperforms single-agent \ac{DQN} and other multi-agent baselines using feedforward networks and the optimization method CPLEX in terms of accuracy and speed. Case studies validate this approach on IEEE 123 and 8500 node test systems \cite{zhao_learning_2022}.

\subsection{Other Use Cases}
Two other use cases for distribution grids include loss minimization and economic dispatch. Loss minimization focuses on optimizing the power grid state by adjusting the topology or the generator set points to reduce losses in branch elements. Economic dispatch minimizes operational generator costs. 
\paragraph{\normalsize{Reinforcement Learning Approach}}
\cite{jacob2022reconfiguring} minimize the \ac{DNR} loss and improve resilience by re-configuring the grid topology via sectionalizing and tie line switches. The states include the grid topology and other grid properties such as power demands and branch currents. The rewards penalize disconnections or radial structure changes and reward loss reduction and feasible network exploration.

Meanwhile, \cite{chen_scalable_2023} focuses on economic dispatch in systems with high renewable energy, optimizing (renewable) generation and \ac{ESS} power under dynamic conditions. The states include load demands, generator outputs and \ac{ESS} state of charge. To optimize economic costs and improve system stability, the rewards are based on voltage violations and balancing economic operation costs and stability.
\paragraph{\normalsize Graph Embedding}
The graph representations in the presented approaches denote substations and buses as nodes, while the edges represent power lines and transformers. Node features typically consist of properties such as load demand, generator outputs, \ac{ESS} state, and time step information. Similar to \cite{jacob2024real}, \cite{jacob2022reconfiguring} employ a \ac{GCAPCN} (cf. Sec. \ref{chap:basics}) to capture local and global features that are used as input to the policy network. The value network of the \ac{PPO} algorithm is a feedforward neural network. In contrast, \cite{chen_scalable_2023} apply the \ac{SAC} algorithm with the \ac{GCN} layers in both the actor and critic to optimize dispatch policies off-policy.










\paragraph{\normalsize Evaluation}
\cite{jacob2022reconfiguring} evaluate their GCAPS-RL method against a feed-forward approach on modified IEEE 13 and 34-bus systems as well as two conventional baseline methods, \ac{MISOCP} and \ac{BPSO}. GCAPS-RL shows superior real-time decision-making and adherence to topological constraints and outperforms the feed-forward counterpart. \cite{chen_scalable_2023} conduct case studies on a modified IEEE39 system with conventional generators, renewable sources, and an \ac{ESS}. Their \ac{GRL} approach outperforms the \ac{OSPI}, \ac{MPC}, and feed-forward \ac{SAC} policies and shows strong convergence, effective policy performance, and superior scalability with cost reductions.

\subsection{Discussion}

Voltage and grid control in emergencies using \ac{DRL} techniques involve several considerations. Reward functions typically address voltage deviations and may also integrate additional factors such as renewables or power loss. However, balancing these objectives poses a non-trivial challenge as mentioned in Sec.~\ref{sec:TG_discussion}.

The control devices considered are mainly PV inverters, \ac{ESS}, and generator power adjustments, with one notable approach that also includes topological actions. Multi-agent setups suit zonal or microgrid distribution networks, as agents can represent different zones and the conducted experiments show that \acp{GNN} improve the robustness in these scenarios. The strategies of handling global system knowledge varies between the approaches, using either global training with local evaluation or centralized critics. \acp{GNN} integrate information from different agents as shown in \cite{mu2023graph}.  

The graph representations used for distribution grid applications are consistent, however evidence from transmission grid literature shows that the choice of representation affects performance and stability. This underscores the need for more research on suitable representations for distribution grids. A start would be the evaluation of the provably meaningful representations used for transmission grid control. In contrast to the graph representations, \ac{GNN} architectures vary significantly. Firstly, the role of the \acp{GNN} in the \ac{RL} algorithms varies widely, with no consensus on their use in the actor, critic, or both.
The node embeddings are mapped to action vectors through various readout methods, including neural networks, autoencoders, and 1D convolutions. Some
approaches share weights between surrogate models
and actors/critics.
With respect to the architectures, Spectral \acp{GCN}, though less common in other domains, are more prevalent here probably due to their roots in signal processing, aligning closely with electrical engineering. Further they are crucial for noise mitigation and filtering, as they function as low-pass filters that suppress noise. They are a particularly strong choice when designing custom, problem-specific filters—such as a Custom Graph Shift Operator (GSO) used to map the relationship between the voltage angle and magnitude of the AC power flow equations—and often achieve high performance in specialized tasks.

As already mentioned in Sec.~\ref{chap:TG}, \ac{GAT} is frequently favored for when robustness is more important that scalability (general control and voltage regulation). For achieving scalability and generalization across diverse DG topologies (e.g., in emergency load shedding), GraphSAGE is the preferred choice. The fundamental Graph Convolutional Network (GCN) often acts as a reliable baseline encoder, particularly in multi-agent setups.

Temporal information is used in some approaches, but most approaches only consider static data. The benefit of using temporal data remains an open question, as it typically increases the complexity of the models.  
Although state-of-the-art architectures like ARMA Networks and TAGNet have demonstrated superior performance in solving power flow and stability tasks (as detailed in Sec.~\ref{gnn_architectures}), they have not yet been integrated into GRL approaches for distribution grids, presenting a highly promising avenue for future research. 
 
Similar to the conclusions in Sec.~\ref{sec:TG_discussion}, \ac{GRL} agents outperform \ac{DRL} agents with fully connected neural networks in transferability and adaptability to topology changes, handling experiments with deleted grid elements or different structures without significant performance drops. The \acp{GNN}' ability to manage noisy or missing data is a considerable advantage, demonstrating robustness in experiments with generator failures, deleted lines, or nodes. This is particularly advantageous given the prevalence of faulty sensor data in real grid operations.

The experiments and evaluations of these approaches cover a wide range of considerations. Most studies conduct experiments on IEEE grids, with grid sizes ranging from small systems with as few as five buses to large networks with up to 8500 nodes. There is a considerable variation in grid size, affecting the approaches' scalability and generalisability. Notably, some methods show effective performance even without access to global information, highlighting the robustness of \ac{GRL} strategies in this context—an essential advantage for real-world applications where data may be unavailable due to technical limitations or privacy concerns, and where local, faster decisions are critical for real-time use. 

Evaluation metrics typically include voltage deviation, network energy loss, or voltage violation rates, reflecting the overarching goal of grid stability. Traditional optimization techniques and heuristics serve as benchmarks in many studies, as these are typically used in practice. The comparison highlights the comparative performance and efficiency gains of \ac{DRL}-based approaches. Benchmarking against other \ac{DRL} methods, including dense-based and CNN-based approaches, sheds light on the advantages of graph-based methods. The experiments highlight the stability and real-time performance of the proposed frameworks on comparatively small power grids. However, given the substantially larger scale of real-world grids, these methods are not yet practically applicable and can be regarded as preliminary proofs of concept.

It should be noted that none of the studies have been carried out on real data. Most approaches rely on simplifications, such as considering only binary actions for load shedding. This further limits the applicability and highlights the need for experimentation in more realistic scenarios. Nevertheless, the studies confirmed the potential of \ac{GRL} for distribution system use cases. 

\section{Other Applications}
\label{chap:other}
This chapter explores \ac{GRL} approaches for related applications, including new energy markets, communication networks for power grids, and \ac{EV} charging scheduling. We consider only approaches that take into account the underlying power grid structure and constraints. Tab.~\ref{table:other_applications} provides an overview of the \ac{RL} method, action type, GNN architecture, grid size, and overall focus of the approaches analyzed.

\begin{table*}[ht!]
\renewcommand*{\arraystretch}{1.4}
    \centering
        \caption{\textbf{Overview of other relevant \ac{GRL} approaches.}\\ In the left column, \textit{Comm.} stands for \textit{Communication Networks} and in the column \textit{Action}, \textit{CS} refers to charging station.}
        \label{table:other_applications}
\begin{tabular}{|M{0.4cm}||P{2.3cm}|P{2cm}|P{3cm}|P{2cm}|P{3.8cm}|}

\hline
& {Approach} 
& {RL } 
& {Action} 
& {GNN}
& {Focus/ Unique Feature}
\\
\hline
\hline

\multirow{1}{*}{{\begin{turn}{90}Energy Market\,\,\,\,\,\,\,\end{turn}}}   

                                                        &Rokhforoz et al. 2023 \cite{rokhforoz_multi-agent_2023} &Multi-agent Actor-Critic&Pricing&GCN&Traditional electricity market\\
                                                        \cline{2-6}
                                                        & Lee et al . 2022 \cite{lee_reinforcement_2022}&DQN, DRQN, Bi-DRQG, PPO&Buy/sell/hold&GCN, Bi-LSTM&P2P power trading, maximizing integration of renewables\\

\cline{1-6}
\multirow{2}{*}{{\begin{turn}{90}Comm.\,\,\end{turn}}}
                                                        &Islam et al . 2023 \cite{islam_software-defined_2023}&Q-Learning&Routing, setting queue service rate& Spectral GCN& Reduce end-to-end latency and \newline packet loss rate\\

\cline{1-6}

 \multirow{2}{*}{{\begin{turn}{90}EV charging\,\,\,\,\,\,\,\,\,\,\end{turn}}}   & Xu et al. 2022 \cite{xu_real-time_2022}&Double prioritized DQN($\lambda$)&CS recommendation&GAT&Combine RL and Dijkstra for CS recomm. and rounting\\
                                    
\cline{2-6}

                                                        &Xing et al. 2023 \cite{xing_bilevel_2023} &Rainbow DRL&CS recommendation, selection of road segments&GAT&Bi-Level RL for CS recomm. and routing\\

\cline{1-6}

\hline

\end{tabular}

\end{table*}

\subsection{Energy Market}

\ac{GRL} opens new possibilities in the energy market, especially in decentralized bidding or direct trading between entities. Traditionally, bidding strategies are centrally managed, requiring full information on all generation units. This is often computationally infeasible due to privacy concerns and results in large-scale problems. Distributed decision-making, using multi-agent \ac{RL} and \acp{GNN}, has the potential to provide efficient and scalable solutions.

We limit our focus to the context of power grids and review two papers using \ac{GRL} to optimize energy trading strategies considering grid topology. \cite{rokhforoz_multi-agent_2023} focuses on the traditional market where generation units set their prices, and a market operator optimizes bids for the lowest overall cost. In contrast, \cite{lee_reinforcement_2022} explores P2P trading, where individuals trade electricity directly, promoting renewable integration.

\paragraph{\normalsize Approaches.}
\cite{rokhforoz_multi-agent_2023} propose a two-level optimization as follows: first, each unit sets a bidding price; second, the market operator determines the market price such that the overall market cost is minimized and the load demand is satisfied. The aim of each generation unit is to maximize its profit. Accordingly, the rewards are calculated based on the determined market prices. The proposed approach is a multi-agent actor-critic algorithm, with one agent per generation unit. At each time step, the actor network of each generation unit selects an action, i.e. a bidding strategy and  a \ac{GNN} critic network which updates the bidding strategy. As input, it receives a graph consisting of buses as nodes including the respective demand as features.
Experiments on the IEEE 30-bus and 39-bus systems show that the \ac{GNN} approach outperforms the baseline using an MLP-based critic, particularly under varying generation capacities. When tested across different systems, the \ac{GNN} also demonstrated better transfer capability.

In the approach by \cite{lee_reinforcement_2022}, energy is traded directly between prosumers without an intermediary market. The setup includes multiple nanogrids, an information network, and a business network for trading. The proposed \ac{RL} algorithm learns trading strategies to minimize maximum load and maximize renewable integration, including power from discharging \acp{EV}.
The agent's actor and critic are hybrid models combining a \ac{GCN} with a Bi-LSTM to process time series data on prosumer consumption and production. The model inputs include cluster demand, renewable supply, system price, and demand response. The actions are either buy, sell, or hold. The reward is based on a rule-based baseline, and multi-objective optimization includes load shifting.
The authors compare various \ac{RL} methods, including \ac{DQN}, Bi-LSTM, and \ac{PPO}, using a nanogrid with real usage data. The \ac{PPO} \ac{GCN}-Bi-LSTM approach achieves the lowest electricity cost and performs better than other methods, significantly reducing average electricity costs with P2P trading.

\paragraph{\normalsize Discussion.}

Both studies demonstrate that \acp{GNN} are promising for optimizing energy markets, particularly as decentralized approaches gain popularity for computational and privacy reasons. \acp{GNN} allow consideration of neighboring market participants' information without creating large-scale problems, unlike traditional deep learning methods that typically treat participants as independent samples. Experiments show that incorporating information from nearby nodes enhances overall market profit. \acp{GNN} improve both Actor-Critic and Q-Learning \ac{RL} methods by capturing interdependencies missed by MLP-based methods, thereby learning more representative grid embeddings crucial for \ac{RL} decision-making. These findings highlight the potential of \ac{GRL} in energy markets, with more \ac{GRL}-based approaches expected in the future.

\subsection{Power Communication Networks}

Apart from the power transmission itself, modern energy systems also transmit information for monitoring and control, requiring efficient routing in communication networks to avoid critical information loss. Unlike physical power transmission (cf. Sec.~\ref{chap:TG} and Sec.~\ref{chap:DG}), these networks operate on the cyber layer.

The study in \cite{islam_software-defined_2023} addresses packet routing and presents a prioritization strategy including different qualities of service, which is not implemented in practice. They distinguish two types of packets: periodic packets with fixed schedules and emergency packets needing low latency to avoid loss of critical packets. The goal is to reduce overall end-to-end latency and packet loss using software-defined networks that adapt dynamically to grid conditions.

Two Q-Learning \ac{RL} algorithms are trained: one for routing paths and another for queue service rates to minimize congestion. The first agent selects feasible paths, while the second predicts queue service rates at switches. Rewards are based on the difference between switch capacity and queue state in order to accelerate queue emptying.

A \ac{GNN} predicts future grid states to inform the queue service rate agent, though the \ac{GNN} is trained separately from the \ac{RL} agent. The model, using spectral \ac{GCN} with Chebyshev polynomials, is trained on IEEE grid traffic data represented as a graph consisting of switches (nodes) and communication links between them (edges). Experiments on the cyber layers of the IEEE-14 and 39-bus systems show the approach's effectiveness in managing grid communication through message exchanges between devices and control centers.

\subsection{Electric Vehicles Applications}

The rapid growth of electromobility is challenging the grid infrastructure, as it increases electricity demand and introduces variable loads. In this context, \ac{DRL} has been studied for charging management \cite{bayani2022, sadeghianpourhamami2020, li2019, silva2020}, station recommendation \cite{xu_real-time_2022,xing_bilevel_2023}, navigation \cite{xu_real-time_2022, xing_bilevel_2023, xing_graph_2023}, and pricing optimisation \cite{zhang_multi-agent_2022}. These applications optimize the allocation of electricity, the pricing, and the routing of \acp{EV}. We concentrate on those \ac{GRL} approaches that consider the underlying power grid and its constraints.

\paragraph{\normalsize Reinforcement Learning algorithms.}
The study in \cite{xu_real-time_2022} tackles the increasing demands of fast charging stations using a multi-objective \ac{DRL} method to dynamically allocate \acp{EV} to stations based on the interest of \ac{EV} owners, charging stations (CS), traffic nodes (TN), and power grid nodes (PG). The agent recommends CSs and guides \acp{EV} using Dijkstra's algorithm, optimizing waiting times, service balance, traffic congestion, and grid voltage deviation. The recommendation of a CS is fast, but the full process is completed only once the \ac{EV} finishes charging. The double-prioritized \ac{DQN}($\lambda$) method is introduced to address this delay and unpredictability. It integrates $\lambda$-return and experience replay with a small buffer to improve efficiency. During training, high-quality samples are prioritized using an attention mechanism, along with a strategy to regulate boundary actions.

Similarly, \cite{xing_bilevel_2023} present a Bi-Level \ac{GRL} approach for charging and routing in Transportation Electrification Coupled Systems. Using a Rainbow-architecture \ac{DRL} block, the high level agent recommends CSs, while the low level agent selects routes with \ac{DRL}. This bi-level approach addresses credit assignment by selecting charging stations at the high level, focusing on the target CS. The low level agent handles the route selection to that station. The reward considers charging costs, battery loss, time allocation, energy consumption, travel time, and voltage limit penalties, with the high level interacting with charging stations and power grids and the low level with traffic nodes.

\paragraph{\normalsize Graph Embeddings.}

\acp{GNN} leverage the inherent graph structure of transportation systems and power grids. Therefore,  \cite{xu_real-time_2022} design a graph structure based on the physical properties. The CSs connect to TNs, and PGs are based on geographical and power supply relations. A unified expression method with type-specific transformation matrices projects features into a shared space, and \acp{GAT} extract meaningful features. These learned representations are integrated with \ac{EV} features for input to a \ac{DRL} agent.

Similarly, \cite{xing_bilevel_2023} utilize \acp{GAT} and introduce an instantaneous adjacency matrix for connections among \acp{EV}, CSs, TNs, and PGs, with smaller matrices representing different relationships. Node features store energy and information features.

\paragraph{\normalsize Experiments and Evaluation.}

The approach in \cite{xu_real-time_2022} is validated using a power-transportation simulation platform with an IEEE 33-node distribution network and a 25-intersection traffic network. They optimize traffic, user experience, and grid stability, outperforming distance-based methods even with charging station queue limits. Training a user-oriented graph \ac{DQN}($\lambda$) agent shows long-term benefits and improved user experience. Combining \acp{GAT} and \ac{DQN}($\lambda$) improves training and decision-making, though a comparison with MLP-\ac{DQN}($\lambda$) would be required to assess the impact of \acp{GAT}.

On the other hand, \cite{xing_bilevel_2023} evaluate their Bi-Level \ac{GRL} approach using real transportation-electrification data, achieving a 10.08\% cost reduction and 16.45\% time savings for owners. Compared to other \ac{DRL} and traditional methods, their \ac{GRL} approach lowers the average total cost by 8.96\% (distance-based) and 4.73\% (\ac{DRL}), demonstrating its efficiency. They recommend learning \acp{GNN} weights directly with \ac{RL} for better robustness and scalability.

\paragraph{\normalsize Discussion.}

The authors of \cite{xu_real-time_2022} demonstrate the effectiveness of \ac{GRL} in dynamic resource allocation for charging stations, emphasizing real-time responsiveness and multi-stakeholder considerations. Their approach highlights the role of sequential decision-making in balancing objectives across transportation and power networks. In contrast, \cite{xing_bilevel_2023} focuses on efficient charging and routing coordination through \acp{GNN}. Their method aims to reduce charging costs and travel time, demonstrating the potential of \ac{GRL} in real-world scenarios. While both employ \ac{GAT} architectures, other \ac{GNN} architectures and robustness to noisy data should be investigated more closely to confirm the applicability of \ac{GRL} in real transportation and energy infrastructures. While current \ac{GRL} approaches demonstrate strong performance in simulation, real-world deployment remains in an early stage, though pilot projects for EV charging management exist \cite{acn_data}. Key challenges include interfacing with heterogeneous transportation and power systems, processing noisy and incomplete data, and coordinating the joint optimization of mobility scheduling and grid constraints. This is logistically complex due to the need for integrated, cross-domain control systems.

\section{Conclusion and Outlook}
\label{chap:outlook}

\paragraph{Conclusion} This survey provides the first comprehensive analysis of Graph Reinforcement Learning for power grid control, revealing a rapidly growing field that, while promising, remains largely in a proof-of-concept phase. Our primary finding is that the core motivation for Graph Reinforcement Learning is not to achieve optimality, but rather to achieve speed, scalability, and adaptability to uncertainty. It excels at finding high-quality, feasible solutions for complex, non-linear, and combinatorial problems—like real-time topology control—where traditional mathematical solvers are computationally intractable. Our review confirms that Graph Neural Networks consistently outperform non-graph-based neural networks in decision-making for power grids, offering superior generalization to unseen topologies and robustness to noisy or missing data. Architecturally, we identified a clear trend away from monolithic agents toward hierarchical and multi-agent systems, with a preference for attention-based and sampling based Graph Neural Networks to enhance robustness and scalability. While we briefly touched upon zoned grids, the specific application of Graph Reinforcement Learning to microgrids represents a vital future direction. As power systems shift toward decentralized generation, Graph Reinforcement Learning is well-suited to optimize the combinatorial switching required for efficient microgrid islanding and self-healing.

\paragraph{Key Challenges and Future Directions} Despite this promise, Graph Reinforcement Learning is not yet deployable. The most significant barrier is the gap between simulation and reality. The field's reliance on the Grid2Op framework, while crucial for development, means that most studies are validated on small IEEE grids that abstract real-world complexities, such as busbar configurations or full N-1 security. This also introduces reproducibility issues if stochastic seeds and horizons are not standardized. Future work should focus on large-scale, open datasets and address the challenges of handling noisy, real-world data. There is an urgent need for a benchmark framework for the distribution grid, specifically for voltage control, to ensure experiments are comparable across all scenarios.  

To bridge this gap, a two-pronged progress is needed. First, the field requires standardized evaluation protocols—including fixed stochastic seeds, consistent evaluation horizons, and operator-aligned metrics (such as minimizing N-1 load flow)—to ensure reproducible and comparable results. Second, the field needs methodological consolidation. Our review identified numerous isolated innovations, including heterogeneous graph representations and specialized Graph Neural Network architectures, hierarchical Reinforcement Learning algorithm designs, and imitation learning for pre-training, which have been developed in parallel but not yet integrated. Future work should combine these and explore state-of-the-art Reinforcement Learning methods like \textit{Bigger, Better, Faster} or full model-based Graph Reinforcement Learning, which remain largely unexplored.

Finally, agents must be safe, moving beyond simple reward penalties to formal Safe Reinforcement Learning techniques. They must be transparent and handle multi-objective optimization to present operators with a Pareto-front of solutions, not just one black-box answer. Incorporating domain knowledge, such as physics-informed losses and custom graph operators, will be crucial for building this trust. 

Ultimately, even after successfully navigating the challenges of realism, standardization, and trustworthiness, deploying Graph Reinforcement Learning into the actual power grid requires strict adherence to operational and legal constraints. From a regulatory perspective, compliance with obligations such as the EU AI Act is essential and mandates extensive testing, documentation, rigorous auditability, and guaranteed human supervision. In addition, real grids experience continuous distribution shifts caused by fluctuating generation and consumption or grid expansion, demanding reliable shift detection and automated retraining. This requires tight integration with existing operator systems which typically suffer from data latency and synchronization issues. Consequently, control agents must be embedded in a robust MLOps pipeline that supports continuous monitoring, retraining, and dependable interaction with operational technologies. 

Despite these challenges, Graph Reinforcement Learning stands out as a promising solution, providing the essential scalability and adaptability required to operate future renewable-dominated power grids.

\section{Acknowledgements}
This work is supported by
\begin{enumerate}
\item Graph Neural Networks for Grid Control (GNN4GC) founded by the Federal Ministry for Economic Affairs and Climate Action Germany under the funding code 03EI6117A.
\item AI4REALNET has received funding from the European Union’s Horizon Europe Research and Innovation program under the Grant Agreement No 101119527. Views and opinions expressed are, however, those of the author(s) only and do not necessarily reflect those of the European Union. Neither the European Union nor the granting authority can be held responsible for them.
\item Reinforcement Learning for Cognitive Energy Systems (RL4CES) from the Intelligent Embedded Systems of the University Kassel and Fraunhofer IEE funded by the Ministry of Education and Research Germany (BMBF) under the funding code 01IS22063A.
\item GAIN project funded by the Ministry of Education and Research, Germany (BMBF), under the funding code 01IS20047A, according to the 'Policy for the funding of female junior researchers in Artificial Intelligence'.
\end{enumerate}
\begin{acronym}[myacronyms]
    \acro{GNN}{Graph Neural Network}
    \acro{GCN}{Graph Convolutional Network}
    \acro{GAT}{Graph Attention Network}
    \acro{EV}{Electric Vehicle}
    \acro{RL}{Reinforcement Learning}
    \acro{DRL}{Deep Reinforcement Learning}
    \acro{GRL}{Graph Reinforcement Learning}
    \acro{XAI}{Explainable AI}
    \acro{TSO}{Transmission System Operator}
    \acro{DSO}{Distribution System Operator}
    \acro{PPO}{Proximal Policy Optimization}
    \acro{CPES}{cyber-physical energy system}
    \acro{OPF}{Optimal Power Flow}
    \acro{DQN}{Deep Q-Network}
    \acro{DDQN}{Double Deep Q-Network}
    \acro{CNN}{Convolutional Neural Networks}
    \acro{A3C}{Asynchronous Advantage Actor-Critic}
    \acro{DDPG}{Deep Deterministic Policy Gradient}
    \acro{DPG}{Deterministic Policy Gradient}
    \acro{PPO}{Proximal Policy Optimization}
    \acro{SAC}{Soft Actor-Critic}
    \acro{TRPO}{Trust Region Policy Optimization}
    \acro{MCTS}{Monte Carlo Tree Search}
    \acro{SACD}{Soft Actor-Critic Discrete}
    \acro{STGCN}{Spatio-Temporal Graph Convolutional Network}
    \acro{BESS}{Battery Energy Storage Systems}
    \acro{SMAAC}{Semi-Markov Afterstate Actor-Critic}
    \acro{ESS}{Energy Storage Systems}
    \acro{GSO}{Graph Structure Operator}
    \acro{MISOCP}{Mixed-integer Second-order Conic Programming}
    \acro{BPSO}{binary particle swarm optimization}
    \acro{GCAPCN}{Capsule-based Graph Convolutional Network}
    \acro{OSPI}{Optimal Solution with Perfect Information}
    \acro{MPC}{Model Predictive Control }
    \acro{RES}{Renewable Energy Sources}
    \acro{DNR}{Distribution Network Reconfiguration}
    \acro{MARL}{Muli-Agent Reinforcement Learning}
    \acro{MDP}{Markov Decision Process}
    \acro{BBF}{Bigger, Better, Faster}

\end{acronym}
\bibliographystyle{splncs04}
\bibliography{08_references}

\end{document}